\documentclass[a4paper]{article}

\usepackage[english]{babel}
\usepackage[utf8]{inputenc}
\usepackage[T1]{fontenc}
\usepackage{lmodern,textcomp}
\usepackage[a4paper,top=3cm,bottom=2cm,left=3cm,right=3cm,marginparwidth=1.75cm]{geometry}

\usepackage{amsmath}
\usepackage{amssymb}
\usepackage{graphicx}
\usepackage{subfig}
\usepackage{float}
\graphicspath{{./figure/}}
\usepackage{multirow}
\usepackage[bottom]{footmisc}
\usepackage{enumitem,cfr-lm}
\usepackage[referable]{threeparttablex}
\renewlist{tablenotes}{enumerate}{1}
\makeatletter
\setlist[tablenotes]{label=\tnote{\arabic*},ref=\arabic*,itemsep=\z@,topsep=\z@skip,partopsep=\z@skip,parsep=\z@,itemindent=\z@,labelindent=\tabcolsep,labelsep=.2em,leftmargin=*,align=left,before={\footnotesize}}
\makeatother
\usepackage[normalem]{ ulem }
\usepackage{soul}
\usepackage[super]{nth}
\usepackage[numbers,square]{natbib}

\usepackage{booktabs}
\usepackage{multirow}

\usepackage{authblk}

\newcommand{\WFscale}{0.7}

\providecommand{\keywords}[1]
{
	\small	
	\textbf{\textit{Keywords---}} #1
}

\usepackage[pagewise]{lineno}

\usepackage[colorinlistoftodos]{todonotes}
\usepackage[colorlinks=true, allcolors=blue]{hyperref}

\title{Unsupervised algorithm for disaggregating low-sampling-rate electricity consumption of households}

\author[1]{Jordan Holweger}
\author[1]{Marina Dorokhova}
\author[1]{Lionel Bloch}
\author[1,2]{Christophe Ballif}
\author[1]{Nicolas Wyrsch}
\affil[1]{École Polytechnique Fédérale de Lausanne (EPFL), Institute of Microengineering (IMT),
Photovoltaics and Thin Film Electronics Laboratory (PV-Lab), Rue de la Maladière 71b, 2000 Neuchâtel, Switzerland}
\affil[2]{CSEM, PV-center, Jaquet-Droz 1, 2000 Neuchâtel, Switzerland}

\begin{document}

\maketitle

\begin{center}
    \vspace{-1em}
    Jordan Holweger\\
    Tel : +41 21 693 42 25\\
    \href{mailto:jordan.holweger@epfl.ch}{jordan.holweger@epfl.ch}\\
\end{center}

\begin{abstract}
Non-intrusive load monitoring (NILM) has been extensively researched over the last decade. The objective of NILM is to identify the power consumption of individual appliances and to detect when particular devices are on or off from measuring the power consumption of an entire house. This information allows households to receive customized advice on how to better manage their electrical consumption. In this paper, we present an alternative NILM method that breaks down the aggregated power signal into categories of appliances. The ultimate goal is to use this approach for demand-side management to estimate \textbf{potential flexibility within the electricity consumption of households}. Our method is implemented as an algorithm combining NILM and load profile simulation. This algorithm, based on a Markov model, allocates an activity chain to each inhabitant of the household, deduces from the whole-house power measurement and statistical data the appliance usage, generate the power profile accordingly and finally returns the share of energy consumed by each appliance category over time. To analyze its performance, the algorithm was benchmarked against several state-of-the-art NILM algorithms and tested on three public datasets. The proposed algorithm is unsupervised; hence it does not require any labeled data, which are expensive to acquire. Although better performance is shown for the supervised algorithms, our proposed unsupervised algorithm achieves a similar range of uncertainty while saving on the cost of acquiring labeled data. Additionally, our method requires lower computational power compared to most of the tested NILM algorithms. It was designed for low-sampling-rate power measurement (every 15 min), which corresponds to the frequency range of most common smart meters.

\end{abstract}
\section*{Acronyms}

\begin{tabular}{rl}
CO & Combinatorial optimization\\
DDSC & Discriminative disaggregation via sparse coding\\
DUE & Device usage estimation\\
FHMM & Factorial hidden Markov model\\
GSP & Graph signal processing\\
HMM & Hidden Markov model\\
ICT & Information and communication technologies\\
NILM & Non-intrusive load monitoring\\
NILMTK & Non-intrusive Load Monitoring Toolkit\\ 
\end{tabular}


\clearpage
\section{Introduction}
The modern energy system differs significantly from conventional energy systems of the past, facing new challenges such as increased self-consumption, aging grids and balancing load and generation in the context of high penetration of renewable energy sources \cite{Strbac2008}. Demand-side management---one of the possible solutions to these challenges---deals efficiently with bidirectional energy flows. It is important for utilities to keep track of users' potential for  flexibility, in order to understand their ability to provide ancillary services to the grid. This can be done using non-intrusive load monitoring (NILM), which can be an instrument to gain insights into users' electrical consumption. An early definition of NILM was given by Hart \cite{Hart1992}: \textquotedblleft A non-intrusive appliance load monitor determines the energy consumption of individual appliances turning on and off in an electric load, based on detailed analysis of the current and voltage of the total load, as measured at the interface to the power source.\textquotedblright \space The underlying concept of Hart's methodology to disaggregate electrical consumption was to analyze the total electric load by measuring current and voltage waveforms. The challenge of NILM is to retrieve the power signals of individual appliances from the total aggregated power signal measured at a single entry point. Mathematically, the problem can be formulated as 

\begin{equation}
\label{eq:Problem}
\boxed{
	\begin{aligned}
	&\boldsymbol{P}^t = \sum_{m} P_{m}^t  +\epsilon_t\\
	&\text{Problem: find an estimate }\hat{P}_m^t \text{ of }P_{m}^t 
	\end{aligned}
}
\end{equation}

\noindent where $\boldsymbol{P}^t$ is the aggregated power signal of a house over time $t$, $P_{m}^t$ 
corresponds to the power signal of the m$^{th}$ appliance and $\epsilon_t$ is the measurement noise. 

Since the introduction of Hart's methodology, the research community has proposed different approaches to disaggregation. As their number continues to increase, the need to classify the approaches emerged. In order to efficiently  match the proposed solutions with specific problems according to their characteristics, researchers have classified the algorithms into several categories: high and low frequency, supervised and unsupervised, residential and industrial, and others.

One categorization technique is based on the frequency of the aggregated power measurements. As proposed in \cite{Esa2016}, low-frequency measurements correspond to sampling rates of 1 Hz and lower, while high-frequency measurements require data of typically a few kHz to a half a MHz. The principle is to measure voltage and current at sufficiently high sampling rates and identify the signatures of individual appliances' loads as in \cite{Liang2010}. These methods are usually based on transient analysis of a power signal, i.e. on extracting the shape and length of a transient waveform. Although high-frequency approaches are highly promising, it is currently not cost-effective to demand such sensing capabilities and data transmission requirements from smart meters \cite{Wang2018}. 

In contrast, methods requiring a significantly lower sampling frequency are typically based on steady-state features such as measuring the instantaneous power signal at a low sampling rate, as described by Zoha et al. \cite{Zoha2012}. Nowadays, most smart meters transmit signals in intervals of 5, 10 or 15 minutes or longer. In an extreme case of divining appliances' consumption with little information, Birt et al. \cite{Birt2012} attempt to disaggregate the power signal received from smart meters at a one-hour sampling rate. They create a regression model on the external and internal temperatures to separate the consumption of heating and cooling systems from the rest of the load. Zhao et al. \cite{1hourStankovic} tackle disaggregation problem on both 15-min and 1-hour electrical measurements by the means of supervised K Nearest Neighbours algorithm. Researchers extract features from time of use profiles of particular appliances and propose a method to select the most useful features per device. Validation on three publicly available datasets has shown the ability of such algorithm to disaggregate up to 62\% of the daily energy consumption.  In \cite{Batra2016}, the proposed methodology uses even less information, since only monthly bills are used to disaggregate the end-use energy consumption into categories (e.g. fridge, lights, washing machine, etc.). Batra et al. do this by relating, using carefully selected features, a household equipped with a single smart meter system to a set of $K$ neighboring households equipped with sub-meters at the appliance level. The consumption of a particular appliance in a test household is predicted by averaging the consumption of the corresponding $K$ sub-metered appliances. 
In the following, we slightly redefine the notion of \cite{Zoha2012} of low and high-frequency measurements. Thus, the prior refers to  sampling rates when the time between two consecutive samples is greater than 5 minutes, while the latter - when it is lower. 
Another way to categorize disaggregation algorithms is to split them into supervised and unsupervised as suggested in \cite{Faustine2017}. In a supervised method, the disaggregation of an unknown power signal is preceded by a training phase, where the algorithm ''learns'' to recognize individual appliances' power signals from the aggregated signal based on available labeled data. An unsupervised method does not require such preliminary treatment as it deals with unlabeled data and can directly perform the disaggregation. In both cases, the approaches were adopted from the field of machine learning. \par
One of the 'classic' supervised techniques of this field is to build an artificial neural network to train an algorithm to create an estimate \cite{Wang2003}. Neural networks have been used successfully in \cite{Ruzzelli2010,kelly_neural_2015,Biansoongnern2016}. An event-based method is another approach to disaggregate real-power measurements, but requires a fairly high level of granularity (typically with a sampling frequency in the range of seconds). A method based on a decision tree was proposed by Liao et al. \cite{Liao2014} and was used in \cite{Stankovic2016} to disaggregate 8s real-power measurements. Similar to computer vision and image processing techniques \cite{Mairal}, dictionary learning through sparse coding was proposed by \cite{Kolter2010} and was tested on a low-sampling-frequency dataset (one-hour time interval). An upgraded version using powerlets as words for dictionary learning was developed in \cite{Elhamifar2015}. A more recent approach \cite{Singh2017} extends the  discriminative sparse coding method by decomposing a problem into multiple sub-problems. A similar technique using extended formulation is the Sum-to-k constrained non-negative matrix factorization (S2K-NMF), tested by \cite{Rahimpour2017}. This method claims the advantage of enabling whole-building disaggregation at a low sampling frequency.

Unsupervised methods, instead, are not concerned with extracting functional dependencies between the data and target variables. Therefore, the preceding training stage of learning from labeled data is absent and algorithms are directly applied to the dataset of interest. A review of unsupervised methods for load disaggregation can be found in \cite{Bonfigli2015}. One common unsupervised method is the factorial hidden Markov model (FHMM) discussed in \cite{Zoha2013} and \cite{Kim}. This method can also be applied in a supervised way as in \cite{Batra2014a,Bonfigli2017}. The most important drawback of most unsupervised methods is their requirement of a relatively high sampling frequency (typically greater than 1/60 Hz). This makes it difficult and costly to collect the necessary data for algorithms' inputs, thus creating obstacles towards implementing such solutions in the real world. 

All the aforementioned methods basically perform analysis and transformation of the power signal. An alternative or complementary approach is to investigate the underlying graph structure of the power signal \cite{Kumar2017a}. This is typically described as graph signal processing (GSP). Successful implementation of such algorithms for disaggregation was performed in \cite{Kumar2017a,He2016,Kumar2016,Kumar2017}. In \cite{Zhao2016}, a GSP approach was used to perform unsupervised disaggregation.

The drawback of supervised methods is the need for labeled data, which are often not public, or not generalizable. Therefore unsupervised methods are more appropriate for general study. The unsupervised methods presented in this paper require either high sampling frequency measurements (smaller or equal to 5 minutes) or do not allow to disaggregate several categories of appliances but rather one versus the rest such as in \cite{Birt2012}. Therefore there is a need for an unsupervised disaggregation algorithm designed for a low sampling rate. 

In this paper, we propose an unsupervised methodology to estimate a household's energy consumption for selected categories based on its characteristics and active power measurements at a low sampling rate (15 min). The method relies only on general information about households and measurements of their energy consumption. Although our proposed methodology does not yet consider space heating or cooling (as the latter is not common in central Europe), it tries to reach a finer level of detail by splitting the remaining load into categories. The categories are formed by grouping appliances together according to their most common activity. Similar to the work of Stankovic et al. \cite{Stankovic2016}, the method allows us to estimate the energy consumed per category of appliances. However, the key difference is that Stankovic used a mix of a supervised NILM method and some individual appliance measurements to investigate the relationship between energy usage and a household's activity, while our methodology is unsupervised. 

The novelty of our methodology consists of proposing a hybrid approach, which lies between load simulation and load disaggregation.  Additionally, we use general household data that was not previously taken into consideration for NILM purposes. Table \ref{tab:hh_ch} summarizes the input features of selected reference algorithms and highlights the original input features required by our algorithm - inhabitants' age groups and employment status, house heating system and appliance usage frequency. This data is easy to acquire as it is commonly used in various occasions either for social science experiments or for customer-service purposes. Hence, this method is less privacy intrusive as its required data granularity is much lower than most common NILM algorithms and the household metadata is generic enough to preserve anonymity. We also generate activity chains for each member of the household, which to our knowledge, was not previously used in any other methods. Therefore, our methodology presents an unsupervised person-centric load disaggregation algorithm. This method cannot be confused with a generalized model of household energy consumption as it relies on both the individual metadata of the households (namely its characteristics) and on the actual measurements of their power consumption. To the best of our knowledge, it is the first alternative, statistical based NILM method suitable for very low sampling rates. All the other unsupervised algorithms require sampling rates equal to 1 min or below, except the method demonstrated in \cite{Birt2012}, where only the heating demand is disaggregated from the rest of the house consumption.

We use three datasets collected in central Europe to test our proposed algorithm and benchmark it against several state-of-the-art NILM algorithms. Additionally, we propose specific performance metrics to compare the algorithms.
This paper is organized as follows: section \ref{sec:Meth} describes the proposed methodology and the approach to benchmark this algorithm against several NILM methods using a selection of public datasets. Section \ref{sec:Results} introduces the results showing how the proposed method performs with respect to the benchmark methods. Finally, section \ref{sec:Conclusion} draws some conclusions and presents possible future work.

\begin{table}[H]
	\centering
	\caption{\label{tab:hh_ch} \bf Summary of the input features for the selected references}
	\begin{threeparttable}
    \begin{tabular}{@{}p{1.8cm}p{1cm}p{1.2cm}p{10cm}@{}}
    \toprule
	\textbf{Ref.} & \textbf{S/U}\tnote{a} & $\Delta t$ & \textbf{Input features}\tnote{1} \\
    \midrule
	\cite{Liang2010}           	    & S\tnote{b} & 1s &  $P$, $Q$, harmonics, instantaneous admittance waveform, current waveform, instantaneous power waveform, eigenvalues, switching transient waveform 	\\			
	\cite{1hourStankovic}   		& S & 1h            & $P$, $Q$, appliance list  \\
	\cite{Batra2016}      			& S & 1month\tnote{c}& $P$, house area, \#rooms, \#occupants, temperature \\
	\cite{Ruzzelli2010}				& S & 1min          & $P$, RMS current,  RMS voltage,  Peak current, peak voltage, sampling rate, power factor, state \\
	\cite{kelly_neural_2015}		& S & 6s            & $P$, appliance power time series of pre-defined window length for model training \\
	\cite{	Biansoongnern2016}		& S & 1s            & $P$, $Q$ \\
	\cite{Liao2014}					& S & 1s-1min       & $P$ \\
	\cite{Kolter2010}				& S & 1h            & $P$ \\				
	\cite{Elhamifar2015}			& S & 1s            & $P$  \\				
	\cite{Singh2017}				& S & 10min         & $P$  \\		
    \cite{Rahimpour2017}			& S & 30s           & $P$  \\	
    \cite{Zoha2013}       			& S & 3s            & $P$, $Q$, appliance list \\
	\cite{Bonfigli2017}				& S & 1min          & $P$, $Q$, individual appliance power signal for model training  \\			
	\cite{Kumar2017}                & S & 3s            & $P$  \\
	\cite{Birt2012}       			& U & 1h            & $P$, temperature, multiple-linear fit\tnote{2} \\
	\cite{Kim}            			& U & 3s            & $P$, appliance list\tnote{2}, states distribution model\tnote{2},  power distribution model\tnote{2}\\
	\cite{Zhao2016}                 & U & 1min          & $P$, database of appliance signatures\tnote{2} \\
    \midrule
    \textbf{Proposed algorithm}		& U & 15min & $P$, house heating system type (electrical or not), appliance list and usage frequency, inhabitants number and age, nominal power per appliances\tnote{2}, activity probability\tnote{2} \\
    	\bottomrule       
    \end{tabular}
    \begin{tablenotes}
        \item[a] S: supervised, U: unsupervised
        \item[b] the proposed method could be applied also in unsupervised manner
        \item[c] disaggregation done at a one-month resolution and training at 15min
        \item[1] for supervised algorithms, input features are used for both training of the model and disaggregation. All of them require the appliance power time series.
        \item[2]  input parameters of the unsupervised models
    \end{tablenotes}
    \end{threeparttable}
\end{table}

\section{Methodology}

\label{sec:Meth}
\begin{figure}[H]
	\centering	\includegraphics[width=0.7\columnwidth]{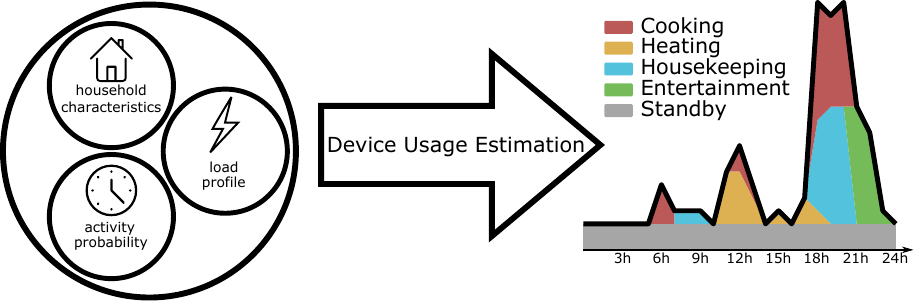}
    \caption{\label{fig:in_algo_out} \bf Graphical representation of the process flow with inputs and output.}
\end{figure}

In this section, we introduce our hybrid approach which we called  the \textit{device usage estimation} (DUE) algorithm. This approach requires  three main inputs: a generic time-of-use survey \cite{SociaalenCultureelPlanbureau2005} from which activities probability are extracted, the household load profile and the household's characteristics which are described in table \ref{tab:hh_ch}. These can be obtained with a survey. Our method consists of using a Markov model to generate activity chains for each person over the age of 10 in the household. Children below the age of 10 are considered to have negligible energy needs, hence they are neglected. Each person's energy consumption is then simulated based on the measured aggregated load curve.  In other words, knowing a household's characteristics and the probability of given activities with time, this algorithm aims at recognizing which appliances are on at which times of the day as shown in figure \ref{fig:in_algo_out}. This problem is pretty close to a disaggregation problem as widely discussed in the literature. As the aggregated load curve comes from smart meters with a sampling rate  lower than what is common in the NILM literature, our approach does not focus on inferring the exact power signal of each appliance at each point in time, but rather on estimating the power signal of categories of appliances, which have been grouped by likely of use. Eight categories of appliances are defined: Cooking,  Entertainment, Fridge, Heating, Housekeeping, Information and communication technology (ICT), Light and Standby. Some additional categories could be added, but the choice of these eight categories comes with the idea of decreasing the number of sub-signals that have to be extracted and improving accuracy. The current implementation of the methodology does not allow a user to define his categories, but it might be possible in future updates.
The complete list of the considered appliances and their category attribution is presented in table \ref{tab:AppData}.

%

\subsection{Algorithm}
The following sections will present the main steps of the methodology, starting with a refresher on Markov models.
A Markov chain describes a stochastic process in which a system may have multiple states $x \in S$. The probability of switching from  state $S_i$ to $S_j$ is described as $a_{i,j}=p(x^{t}=S_i|x^{t-1}=S_j)$. The resulting matrix $a_{i,j}$ is called the transition matrix. The initial condition for the first element of the chain is described by the initial probability distribution $\pi_i=p(x^0=S_i)$.
A hidden Markov model (HMM) states that the system is observed through a set of external variables $y$ that can have multiple states $O_k$ (the external variable can also be continuous) linked to the system states by the relation $b_{k,i}=p(y^t=O_k|x^t=S_i)$. This matrix is called the emission matrix \cite{Rabiner1989a}.

In the proposed methodology, an activity chain is modeled as a Markov process. The transition matrix and initial probability distribution also depend on two main external features. One is related to the type of  day $D \in$ [weekday, Saturday, or Sunday]; the other is related to the household's characteristics, namely,  the employment state $E \in$ [full-time, part-time, student, retired, unemployed] and the age group $G \in$ [teenager, adult active, senior active, senior inactive] of the inhabitants. Hence the transition matrix and initial probability distribution are functions of $D$, $E$ and $G$. 

The 2005 Netherlands time-of-use survey \cite{SociaalenCultureelPlanbureau2005} was chosen as a representative data source to extract the activity probability. Although it is questionable if the behavior of Dutch citizens is representative of the behavior of all central Europeans, it seems like a reasonable assumption, and this limitation could be overcome by the use of similar surveys of other countries such as the  2014-2015 UK time-of-use survey \cite{Gershuny2017}. 

A time-of-use survey provides a diary of activities, in which individuals record what activity was done at each time of the day. In the Dutch time-of-use survey, the reported activities were divided into a set of 14 activity states $S$, summarized in table \ref{tab:ActDev}.  From the diaries, it is possible to extract an activity event table in which each event $i$ is described by an activity $s_i \in S$, that is performed by an individual who belongs to an employment group $e_i \in E$ and age group $g_i \in G$, a start time $t^0_{i} \in T$, an end time $t^1_{i} \in T$ and a type of day $d_i \in D$. $T$ is the time discretization of a day (here the day is discretized at a 5-min resolution). 
For each employment group $e^\ast \in E$, age group $g^\ast \in G$ and type of day $d^\ast \in D$, the computation of the initial state (activity) probability distribution is described by equation \ref{eq:initP}: 

\begin{equation}
\label{eq:initP}
	\begin{aligned}
	 \pi_{e^\ast,g^\ast,s^\ast}& = \frac{\sum_i(\delta_{e^\ast,g^\ast,t^0,i} \cdot \delta_{s^\ast,i})}{\sum_i(\delta_{e^\ast,g^\ast,t^0,i})} \quad \forall s^\ast \in S \\
	\text{where }\\
	 \delta_{e^\ast,g^\ast,t^0,i} &= 
	\begin{cases}
	1,& \quad \text{if } e_i=e^\ast \cap g_i=g^\ast \cap t^0_i =\text{00:00}\\
	0,& \quad \text{otherwise}
	\end{cases}\\
	 \delta_{s^\ast,i} T &= 
	\begin{cases}
	1,& \quad \text{if } s_i=s^\ast \\
	0,& \quad \text{otherwise}
	\end{cases}
	\end{aligned}
\end{equation}

To compute the coefficient of the transition matrix, a transition event table is derived from the activity event. Each transition $j$ is defined by two activities $s^0_j,s^1_j\in S$, at time $t_j \in T$ defined such that for a transition from the activity event $k$ to the activity event $l$, $t_j=t^{end}_k=t^{start}_{l}$. As for the activity event table, the employment and age group $e_j$ and $g_j$ is reported. Similar to the initial probability distribution calculation, for each group $e^\ast \in E$, $g^\ast \in G$ and for each type of day $d^\ast \in D$, the coefficient of the transition matrix is computed as described by equation \ref{eq:transP}:

\begin{equation}
\label{eq:transP}
\begin{aligned}
 a_{e^\ast,g^\ast,s^{1\ast},s^{2\ast},t^\ast} & = \frac{\sum_j(\delta_{e^\ast,g^\ast,t^\ast,j} \cdot \delta_{s^{1\ast},s^{2\ast},j})}{\sum_j(\delta_{e^\ast,g^\ast,t^\ast,j} \cdot \delta_{s^{1\ast},j})} \quad \forall s^{1\ast},s^{2\ast} \in S \quad \forall t^\ast \in T \backslash [\text{00:00}] \\
 \text{where }\\
 \delta_{e^\ast,g^\ast,t^\ast,j} &= 
\begin{cases}
1,& \quad \text{if } e_j=e^\ast \cap g_j=g^\ast \cap t_j=t^\ast\\
0,& \quad \text{otherwise}
\end{cases}\\
 \delta_{s^{1\ast},s^{2\ast},j} &= 
\begin{cases}
1,& \quad \text{if } s^1_j=s^{1\ast} \cap s^2_j=s^{2\ast} \\
0,& \quad \text{otherwise}
\end{cases}\\
  \delta_{s^{1\ast},j} &= 
\begin{cases}
1,& \text{if } s^1_j=s^{1\ast} \\
0,& \text{otherwise}
\end{cases}
\end{aligned}
\end{equation}

In order to generate an activity chain (the exact procedure is described in appendix \ref{APP:ActChain}), a duration probability distribution is derived from the activity event table by extracting the mean and standard deviation from the activity duration $t^{end}_i-t^{start}_i$ for each activity $s$, each group $e \in E$, $g \in G$ and each type of day $d \in D$.

This step does not consist of a learning phase, as it does not use any electrical information and is completely generic within the scope of the mentioned assumptions. 
The following will describe how the input parameters $\pi$ and $a$ are used to break-down the power measurement of a house into 8 categories. Each category consists of a group of appliance as presented in table \ref{tab:AppData}. 

\subsubsection{Overview}
\begin{figure}[H]
	\centering
	\includegraphics[scale=\WFscale]{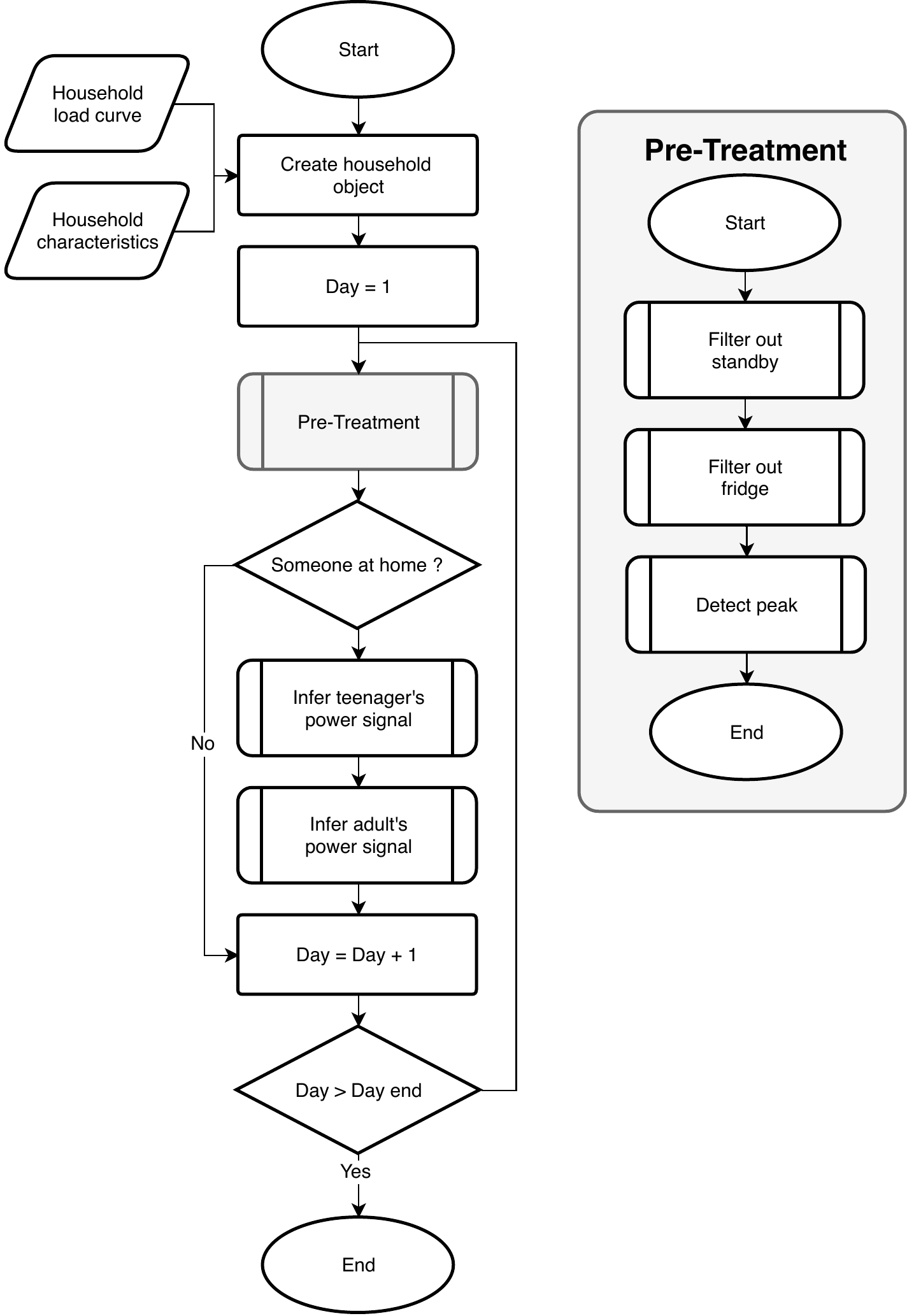}
	\caption{\label{fig:WF_main} \bf General workflow of the proposed methodology. The \textt{pre-treatment} procedure is described in appendix \ref{APP:PreTreatment}.}
\end{figure}

As presented previously, the basic idea is to find an estimate $\hat{P}_m^t$  of $P_{m}^t$ from the total power signal $\boldsymbol{P}^t$ as stated in equation \ref{eq:Problem}. The approach chosen here is to link the observed power signal $\boldsymbol{P}^t$ with the activity chain of each household inhabitant. The main workflow is presented in  figure \ref{fig:WF_main} and can be described as the following. For a given household, knowing its characteristics, and a given measured power signal $L(t)$, the algorithm takes each day separately and performs a sequence of actions:

\begin{itemize}
	\item Filter out the standby consumption by identifying the minimum power over the considered day. $\hat{P}_{standby}^t=\min_t(\boldsymbol{P}^t) \forall t \in T$ where $T$ is the time domain of the considered day. This definition of the standby results in attributing constant part of the fridge consumption to standby. Thus, the successive filtering of the fridge's consumption pattern deals only with the variable part of the fridge consumption.
	\item Filter out the fridge. For the first day, the typical periodic signal (considered over all nights of the load curve) is extracted. This is assumed to be the fridge's consumption pattern. Then for each subsequent day, the fridge signal is synchronized with the measured power signal before filtering out. The seasonal variations of the fridge consumption are taken into account by capturing daily the new signal pattern. 
	\item Detect the peaks using the method from \cite{Billauer2012}.
	\item If any peak indicates that someone is at home for this day, infer the power signal (including lights) generated by each person. This procedure is described in the next paragraph. 
\end{itemize}
More details on how the standby and fridge are filtered out are provided in appendix \ref{APP:PreTreatment}.

\begin{figure}[H]
	\centering
	\includegraphics[scale=\WFscale]{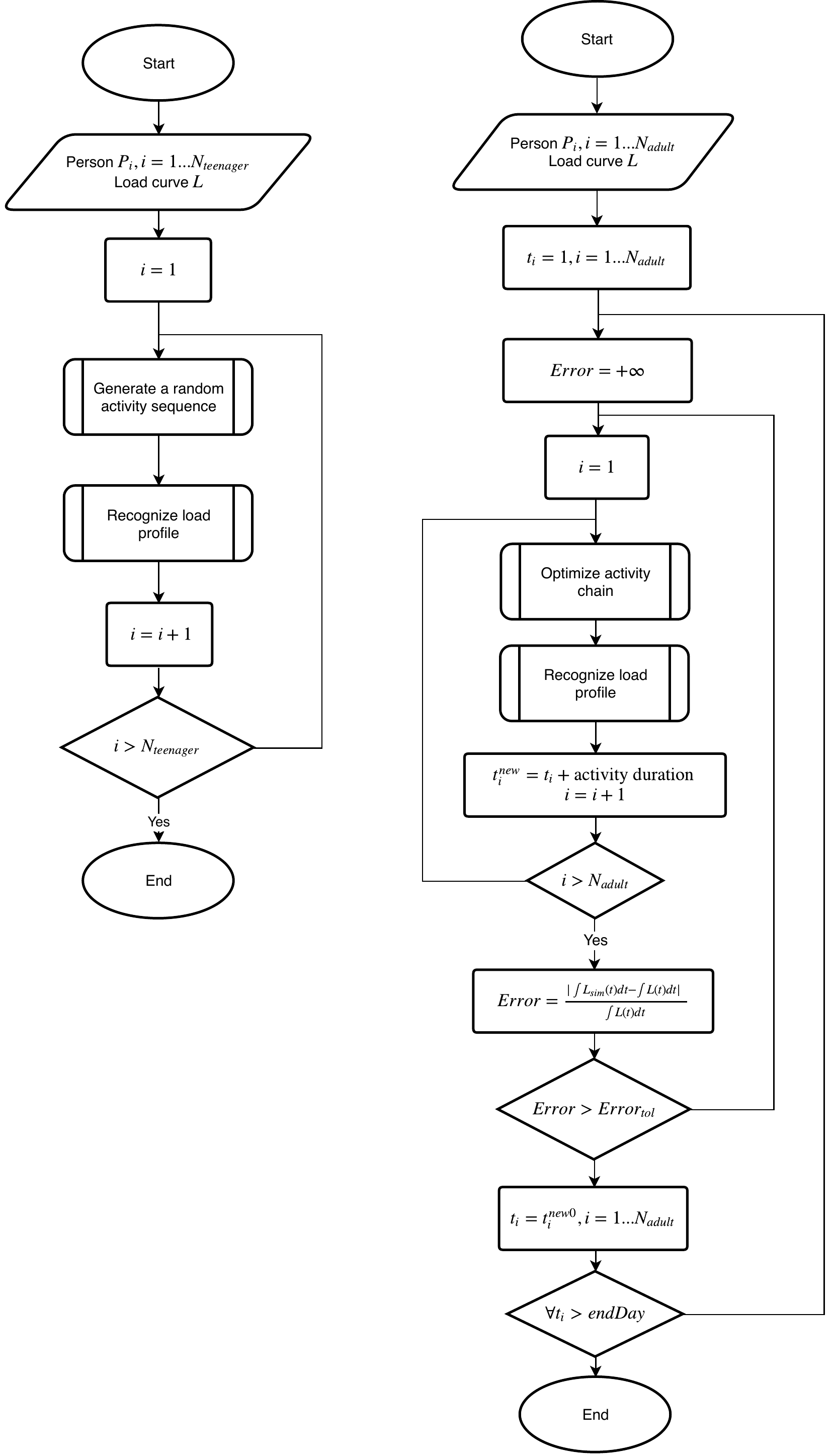}
	\caption{\label{fig:WF_infer} \bf Methodology to infer the individual power signal of a teenager (left) and adult (right).}
\end{figure}

The methodology to infer the power signal that could be generated by each person in the household can be summarized as follows. An activity chain is generated for each person in the household, then a corresponding power signal is estimated. The procedure is repeated until the difference between the measured and simulated load curves is smaller than a given tolerance. The workflow of this process is depicted in figure \ref{fig:WF_infer}.

The inhabitants of the household are treated separately in this step. Teenagers (10 to 18 years old) are considered to have unpredictable activity chains and  are treated separately as depicted in figure \ref{fig:WF_infer}. Adults, the main energy consumers of the households, are considered together in the optimization step. If any power peaks are detected in the considered daily load curve, all inhabitants are assumed to be at home. No partial occupancy is considered here, although this could be a path for further improvements.   

The following section will give some details about the most important subprocesses in figure \ref{fig:WF_infer}. In particular the ''optimize activity chain'' step and the ''recognize load profile'' step will be detailed. The ''pre-treatment'' procedure and ''recognize load profile'' step are described in appendix \ref{APP:PreTreatment} and \ref{APP:Recog} respectively. 

\subsubsection{The ''Optimize activity chain'' step}

This step aims at identifying a possible activity for a given household inhabitant considering the available energy budget for the respective time frame and the characteristics of this person. 
This process is divided into three steps: 

\begin{enumerate}
\item First, a list of possible activities and their corresponding probability of occurrence are defined. In other words, this step selects transition matrix $a$ (or initial probability distribution $\pi$, if it corresponds to the first element in the chain) according to the type of  day $d$, the employment state of the person $e$ and the person's age group $g$. The whole list of possible activities is summarized in table \ref{tab:ActDev}. 

\item Secondly, an activity is selected based on the probability of this activity occurring (i.e. the transition matrix or the initial probability distribution), as well as the duration of this activity based on the duration probability distribution corresponding to this activity.

\item Finally, the compatibility between the chosen activity and the measured load curve is evaluated. If the measured load power is low (relative to the mean power level of the  inventory of household appliances), the selected activity cannot be an activity that requires high-power appliances and vice versa. This step applies only to adults. For teenagers, it is assumed that the activity chain is random. 

\end{enumerate}
The algorithm that randomly selects activities and durations from a probability distribution is presented in appendix \ref{APP:ActChain}. Our method relates the energy usage to the activities within it by assuming the appliance usage probability per activity and the power demand per device. The activity chain could be used as an additional output of the algorithm but this is not the main scope here and would require a particular caution for the analysis. This is an opposite approach  from the one used by Stankovic et al. \cite{Stankovic2016}, who employed a supervised NILM method and individual appliance monitoring to actually correlate the activity chain from the disaggregated energy consumption. 

\subsubsection{The "Recognize load profile" step}
\begin{figure}[h!]
	\centering
	\includegraphics[scale=\WFscale]{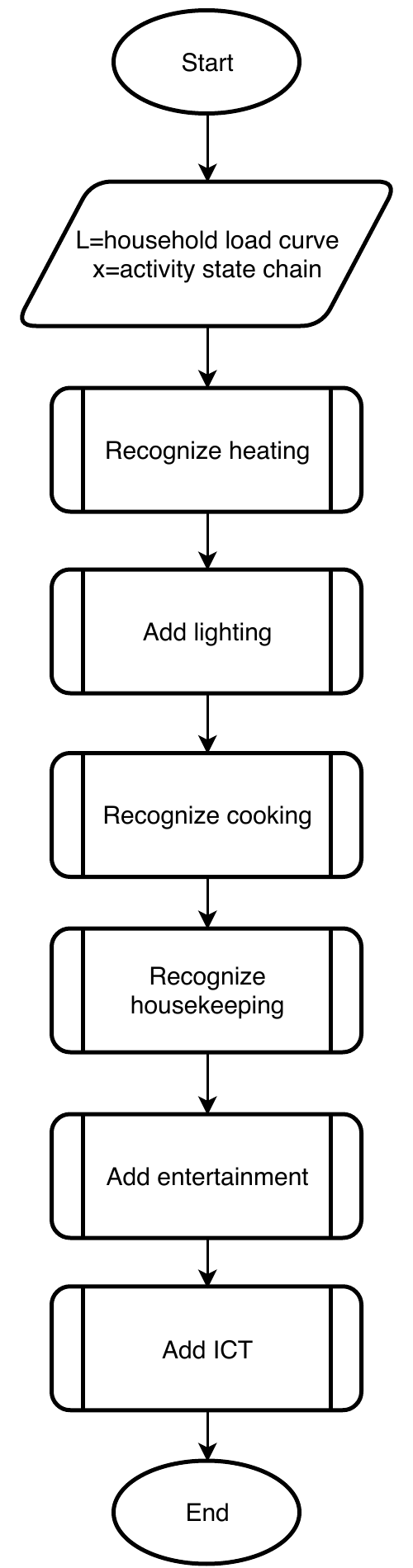}
	\caption{\label{fig:WF_recog_main} \bf Workflow to infer each category of appliance in the aggregated power signal.}
\end{figure}

At this stage, an assumption on the activity of a single inhabitant and the duration of that activity has been formulated. Based on this hypothesis, this step aims to infer the possible power signal of each appliance category sequentially. As shown in figure \ref{fig:WF_recog_main}, the sequence of appliance category recognition is organized so that the most-energy-consuming activities are treated first. The Light category is an exception, as it is treated after the Heating category. The reason for this is that lighting has a weak correlation with the type of activity but  a strong correlation with the occupancy. 

The standard procedure for each of these substeps is the following: 

\begin{itemize}
	\item Identify the time period in which the activities that correspond to the category were detected. 
	\item For each appliance that belongs to this activity and exists in the household inventory, check if the energy budget and time budget are sufficient to run the appliance. Note that at every iteration of this step, the  order of devices is set randomly so that it does not always start with the same appliance. The list of appliances that belong to each activity is presented in table \ref{tab:ActDev}. 
	\item If the energy budget and time budget are sufficient, set the device as used and simulate the corresponding power signal. If the house has several appliances of the same type, a power signal is generated for each appliance. Based on the NILM wiki\footnote{\url{http://wiki.nilm.eu/appliance.html}} and field experience, the nominal power, probability and duration usage for each appliance have been reported in table \ref{tab:AppData}. These reference values should be updated with the appliances' market evolution and introduction of more efficient devices. 
\end{itemize}

\begin{table}[H]
\centering
	\caption{\label{tab:ActDev} \bf List of possible activity states and related possible appliances used.}
	\begin{tabular}{@{}ll@{}}
		\toprule
		\textbf{Activity} & \textbf{Appliances} \\ 
		\midrule 
		Cleaning & vacuum, TV, stereo, lights \\ 
		Using a computer  & TV, stereo, PC, laptop, printer, lights \\ 
		Cooking & stove, oven, microwave, kettle, TV, stereo, lights \\ 
		Washing dishes & dishwasher, TV, stereo, light \\ 
		Eating & coffee maker, microwave, kettle, TV, stereo, lights \\ 
		Do the homework & TV, stereo, PC, printer, laptop, lights \\ 
		Playing a game  & TV, stereo, gaming console, lights \\ 
		Laundry & washing machine, tumble dryer, TV, stereo, lights \\ 
		Music & stereo, PC, tablet, laptop, lights \\  
		Outdoor & $\varnothing$ \\ 
		Sleeping & $\varnothing$ \\ 
		Watching TV & TV, DVD player, PC, tablet, laptop, lights \\ 
		Showering & hairdryer, TV, stereo, lights \\ 
		Working &  $\varnothing$\\ 
		\bottomrule
	\end{tabular} 
\end{table}

\begin{table}[H]
	\centering
	\caption{\label{tab:AppData} \bf Appliance per category and nominal power, parameter of usage and duration.}
	\begin{threeparttable}
	\begin{tabular}{@{}llcccccl@{}}
    \toprule
		\textbf{Category}                            & \textbf{Appliance}       & $\boldsymbol{P_\text{Nominal}}$ \textbf{(W)} & $\boldsymbol{\beta_1}$ &  $\boldsymbol{\beta_2}$ &  $\boldsymbol{\beta_3}$ &  $\boldsymbol{\tau}$ \textbf{(min)}&\textbf{Note} \\
\midrule
		\multirow{5}{*}{Cooking}					 & coffee maker           & 800              & 0.8     & 0.7     & 0.5     & 3	&\tnotex{tnote:r1}        \\
							                         & microwave               & 1250             & 0.3     & 0.5     & 0.4     & 5    &\tnotex{tnote:r1}    \\
													 & kettle                   & 1800             & 0.3     & 0.5     & 0.8     & 2   	&\tnotex{tnote:r1}       \\
													 & oven                     & 2400             & 0.1     & 0.3     & 0.4     & 50   &\tnotex{tnote:r1}       \\
							                         & stove            & 500              & 0.5     & 1.0     & 1.0     & 30   &\tnotex{tnote:r1}        \\
\midrule
		\multirow{8}{*}{Entertainment}              & TV                       & 124              & 0.9     & 0.1     & 0.5     & 20   &\tnotex{tnote:r2}        \\
        											& TV box                   & 20               & 1.0     &         &         &      &\tnotex{tnote:r2}      \\
													 & DVD player              & 80               & 0.1     & 0.0     & 0.0     & 0    &\tnotex{tnote:r2}      \\
													 & PC                       & 110              & 0.5     & 0.1     & 0.2     & 30   &\tnotex{tnote:r4} \ \tnotex{tnote:r5}       \\
                                                     & laptop                   & 55               & 0.5     & 0.2     & 0.4     & 20   &\tnotex{tnote:r4} \ \tnotex{tnote:r5}     \\
													 & tablet                   & 7                &         &         & 0.4     &      &\tnotex{tnote:r4}     \\
													 & stereo                     & 100              & 0.9     & 0.2     & 0.5     & 20   &\tnotex{tnote:r2}      \\
													 & gaming console           & 180              & 0.3     & 0.0     & 0.1     & 80   &\tnotex{tnote:r2}      \\
\midrule
		\multirow{3}{*}{Fridge}                      & fridge (with a freezer)    & 94               & 0.3     & 0.3     &         & 25   &\tnotex{tnote:r6}      \\
        											 & fridge (without a freezer) & 66               & 0.3     & 0.3     &         & 25   &\tnotex{tnote:r6}     \\
													 & freezer alone              & 62               & 0.5     & 0.5     &         & 63   &\tnotex{tnote:r6}     \\
\midrule
		\multirow{3}{*}{Heating}                     & hairdryer               & 600              & 0.2     &         &         &      &        \\
													 & boiler                   & 2000             &         &         &         &      & \tnotex{tnote:r9}         \\
													 & heat-pump                & 1000             &         &         &         &      &\tnotex{tnote:r9}          \\
\midrule
		\multirow{4}{*}{Housekeeping}                & washing machine             & 406              & 0.5     & 0.4     &         & 60   &\tnotex{tnote:r7}         \\
        											 & tumble dryer             & 2500             & 0.5     & 0.0     &         & 60   &\tnotex{tnote:r10}       \\
												  	 & dishwasher              & 1131             & 0.4     & 0.0     &         & 34   &\tnotex{tnote:r7}         \\
													 & vacuum            & 2000             & 0.5     & 0.2     &         & 10   &\tnotex{tnote:r7}       \\
\midrule
		ICT                                          & printer                  & 23               & 0.1     & 0.1     &         & 5    & \tnotex{tnote:r5}     \\
\midrule
		Light                                        & lighting                 & 137              & 0.25     &         &         &      &\tnotex{tnote:r8}     \\
\midrule
		Standby                                      & modem (and similar)           & 8                & 8.0    &         &          &      &\tnotex{tnote:r11}\\
\bottomrule              
	\end{tabular}
	\begin{tablenotes}
		\item\label{tnote:r1}: $\beta_{1,2,3}$ probability of usage for respectively breakfast, lunch, dinner. These values can be set to 0 according to the households habits (namely number of lunch and dinner at home), if provided as presented in table \ref{tab:HsldInfo} from appendix \ref{APP:HsldInfo}.
		\item\label{tnote:r2}: $\beta_1$ probability of usage when the activity is \textit{Watching TV}, $\beta_2$ probability of usage for other activities, $\beta_3$ probability of use if one is already used (for an additional person).
		\item\label{tnote:r4}: $\beta_3$ probability of usage when it is used as a replacement for TV. This is used if no TV is reported in the appliance ownership table as presented in table \ref{tab:HsldApp} from appendix \ref{APP:HsldInfo}.
		\item\label{tnote:r5}: $\beta_1$ probability of usage when the activity is \textit{Using computer}, $\beta_2$ probability of usage when the activity is \textit{Working} or \textit{Do the homework}
		\item\label{tnote:r6}: $\beta_1$ duty cycle during the day, $\beta_2$ duty cycle at night, $\tau$ is the duration of the active cooling phase
		\item\label{tnote:r7}: $\beta_1$ probability of usage once during the day, $\beta_2$ probability of usage additional times. This probability can be set to 0 according to the households habits (namely usage of the washing machine per week) as provided in table \ref{tab:HsldInfo} from Appendix \ref{APP:HsldInfo}.
		\item\label{tnote:r8}: $\beta_1$ $P_{Nominal}$ is used for the first person in the house, $P_{Nominal}*\beta_1$	is used for each additional person.	
		\item\label{tnote:r9}: Presently our algorithms only accounts for hairdryers. 
		\item\label{tnote:r10}: $\beta_1$ probability of usage right after washing machine. This probability can be set to 0 according to the household habits if provided as reported in table \ref{tab:HsldInfo} from Appendix \ref{APP:HsldInfo}.
		\item\label{tnote:r11}: The modem  (or internet router) is  in the Standby category for the obvious reason that it is always on. Its power consumption is assumed to be independent of occupancy and the inhabitants' activities.
	\end{tablenotes}
\end{threeparttable}
\end{table}

Two categories do not follow the standard procedure described above, Light and Housekeeping. Regarding Light, a base power signal is generated from the data of table \ref{tab:AppData}. For each additional person, a new signal is generated and scaled down by a factor of 0.25 so that it does not predict excessive light consumption. We added the 0.25 factor after noticing that our method over-estimated  the light consumption if we assumed the light needs are unshared by the inhabitants.
The Housekeeping's process also differs from the standard procedure. For each appliance in this category (taken in a random order as above), the probability of usage of any particular appliance is set to 0, if such appliance has already been used a number of times during the week, according to the households habits (if provided). The households habits data are described in appendix \ref{APP:HsldInfo}.  \\
For any appliances, the code will search for a period in which the measured power is above the nominal power of the appliance during a sufficient time period (according to the mean appliance usage duration) and generate the corresponding power signal during this time frame. 
At each of these steps, the simulated load curve is subtracted from the measured load curve in order to update the power and energy budget. More details about each individual subprocess are provided in appendix \ref{APP:Recog}.

Our method differs from the methodologies described in the literature in two ways:

\begin{enumerate}
	\item No training is required since the model parameters are extracted from the time-of-use survey \cite{SociaalenCultureelPlanbureau2005}. Appliance power characteristics and usages are extracted from various data sources. 
    
	\item It is a hybrid between a load profile simulator and a disaggregation algorithm. The error on energy in  the main loop is minimized before going to the next time-step, ensuring that the selected solution is as close as possible to the original load profile.
\end{enumerate}

\subsection{Testing}
\label{sec:Test}
In order to assess the validity and performance of our DUE algorithm with respect to other methods of the field, this section will present different NILM algorithms and performance metrics. In order to obtain a confidence interval on the selected metrics, the tests were performed successively on different datasets. 

\subsubsection{Selection of algorithms} \label{subsubsec:algo}
The following set of algorithms was chosen to benchmark the performance of the DUE algorithm: 

\begin{itemize}
\item Combinatorial optimization (CO) \cite{Hart1992}
\item Factorial hidden Markov model (FHMM) \cite{Kim} 
\item Graph signal processing (GSP) \cite{Kumar2017a,He2016}
\item Discriminative disaggregation via sparse coding (DDSC) \cite{Kolter2010}
\end{itemize}

\paragraph{Combinatorial optimization}
\label{sec:comb_opt}
is a well-studied benchmark algorithm presented by Hart \cite{Hart1992}. Assuming a linear model for each appliance $i=1...M$, a Boolean switch process $a(t)=a_i(t),i=1...M$ identifies which of the appliances are "on" and "off" at time $t$. Therefore, the aggregated power load at time $t$ is the sum of all the individual power loads of the appliances that are "on" at that time. Naturally, a combinatorial optimization problem emerges in order to minimize the difference between the predicted and observed power: $\hat{a}(t)=\min_{a_i}|P^t-\sum_{i=1}^n a_iP_i|$ , where $\hat{a}(t)=\hat{a}_i(t)$ is a matrix representing the estimated "on" or "off" state of an appliance in time and $P_i$ represents the nominal power of appliance $i$.\\
The learning phase will construct a power basis $P_i$ from the disaggregated signal. The disaggregation phase will solve the problem for all $a_i(t)$ at every time-step.
This approach is computationally intractable, as the complexity of the algorithm grows exponentially with the number of appliances. Thus one cannot solve the problem exactly. Additional difficulties of this method include detecting simultaneous appliance state changes and a lack of full information about the individual power loads. Therefore, a switch continuity principle was adopted that supposes that only a small number of appliances can change state at the same time. \par
For this study, the CO algorithm was implemented through the Non-Intrusive-Load-Monitoring-Toolkit (NILMTK) \footnote{http://github.com/nilmtk/nilmtk} \cite{Batra2014a}.

\paragraph{The factorial hidden Markov model} is another benchmark algorithm provided in the NILMTK \cite{Batra2014a}. Considering that the only observable value is the aggregated power measurement $\boldsymbol{P}^t$, each appliance's individual power load is modeled separately as a hidden Markov model (HMM) with hidden states $x_i(t)$ representing the status of the appliance. The use of a factorial hidden Markov model (FHMM) over HMM reduces parametric complexity when modeling time series generated by the interaction of several independent processes --- in our case, several appliances \cite{Kim,Kolter}. In order to build the model, four main components need to be defined:

\begin{itemize}
\item Finite set of hidden states
\item Transition matrix $a$, which represents the probability of changing a state
\item Emission matrix $b$, which indicates the probability of emitting an observation
\item Initial distribution of probabilities among the states $\pi$
\end{itemize}

The learning phase consists of building both matrix $a$ and $b$ and the disaggregation phase aims at finding hidden states such that the probability of observing the signal  $\boldsymbol{P}^t$ is maximized. In other words: $\hat{x}_i^t=\max_x p(\boldsymbol{P}^t|x_i^t)$
The DUE algorithm is not a derivation of the FHMM since it relies  on a try-and-fail method rather than on an optimization problem. 

\paragraph{Graph signal processing} is a novel supervised concept for load disaggregation, which is neither state- nor event-based. The success of the methodology for applications such as filtering, clustering, classification, convolution and modulation \cite{Kumar2017a,Kumar2016} inspired researchers to develop a solution to the NILM problem using GSP \cite{He2016,Stankovic2014,Zhao2016,Kumar2017a,Kumar2016,Kumar2017}. The approach relies on the regularization of graph signals, assuming that if the signal is piecewise-smooth, then the total graph variation is generally small. \cite{He2016,Stankovic2014} The proposed GSP methodology has several advantages over conventionally used algorithms: short training periods \cite{He2016}, reliable performance in the presence of noise, unknown or uncommon appliances \cite{Zhao2016}, the ability to handle different sampling rates, including every 15 minutes \cite{Kumar2017a,Kumar2016} and the use of active power alone \cite{Kumar2017}.    \par
The basis of the algorithm lies in constructing an undirected graph $G$ using aggregated power measurements, where each vertex $V$ corresponds to a load sample. The weights $W$ of the edges connecting the vertices reflect the degree of similarity between nodes \cite{Stankovic2014}. The overall sequence of the algorithm's main steps can be summarized as follows in accordance with the formulation proposed in \cite{Kumar2017a,Kumar2016}:

\begin{itemize}
\item Learn the weights $W$ from aggregated power data using a Gaussian kernel weighting function
\item Construct the graph $G$ using $W$ and compute the respective graph Laplacian $L$ 
\item Define graph signals $s$, where $s$ equals the appliances' ground truth (GT) during the training period and equals zeros during the testing period
\item State an optimization problem of minimizing the smoothness term $||s^TLs||_2^2$
\item Evaluate signal $s^*$ as a solution to the optimization problem
\item Subtract the disaggregated signal $s^*$ from the aggregated power measurements
\end{itemize}

The procedure described above has to be repeated sequentially for each appliance, starting from the highest-consuming one. The GSP approach for this study was implemented in Matlab using the Graph Signal Processing Toolbox \cite{perraudin2014gspbox}.

\paragraph{Discriminative disaggregation via sparse coding}
\label{sec:dis_sparse_cod}

Considering that the NILM problem can naturally be formulated as a single-channel source separation problem, it is reasonable to apply methods that are widely used in the field, such as sparse coding. The idea of the algorithm is to train models separately to find approximate representations for each individual appliance in the form of $X=BA$, where $B$ is the set of basis functions, also called a dictionary, and $A$ is a sparse activations matrix. Therefore, the objective function of the method is $\min_A\frac{1}{2}||X-BA||^2_F+\lambda||A||_1$ subject to $A,B\geq0$ The training phase will learn the dictionary $B_i$ for each appliance ($i=1...n$) from the individual appliance signal $X_i$ by  solving the equation  $\min_{A_i}\frac{1}{2}||X_i-B_iA_i||^2_F+\lambda||A_i||_1$ for $A_i$ and $B_i$ sequentially. Then the disaggregation will solve for $A$ using the aggregated signal $X$\cite{Kolter2010,Elhamifar2015,Leijonmarck2015,Yu2016}.\par 
One method to decompose the aggregated signal into a sparse combination of dictionary elements is non-negative sparse coding \cite{Hoyer}. When this approach was utilized for solving the NILM problem, many improvements to the method emerged. Kolter et al. \cite{Kolter2010} proposed to discriminatively optimize basis functions in order to minimize disaggregation error. Elhamifar and Sastry \cite{Elhamifar2015} incorporated additional priors such as device sparsity, knowledge of cooperating devices and temporal smoothness, and Singh et al. \cite{Singh2017} extended the approach to multiple layers of dictionary learning for each device. \par
This method is potentially suited for low-sampling rate datasets as demonstrated in \cite{Kolter2010} an attempt to disaggregate meter readings at 1h interval. 
For the current study, the DDSC algorithm was implemented in Python \cite{Python} based on the representation in \cite{Kolter2010,Leijonmarck2015}.

\subsubsection{Test datasets}
Instead of testing on a single dataset as it is usually done in the literature, we propose a broader experiment by testing on several datasets. The underlying objective is to get a confidence interval on each of the selected performance metrics. Our requirements for the datasets are the following: 
 
\begin{itemize}
	\item Each dataset must consist of households with information about consumption at the appliance level.
	\item Each households must be located in Europe as time-of-use survey \cite{SociaalenCultureelPlanbureau2005} was collected in Europe
	\item There must be a sufficient description of the characteristics of each inhabitant 
\end{itemize}

In order to emulate data acquired from smart meters (considered here as average power over a particular time interval), all datasets were down-sampled to a 15-min sampling rate, keeping the real power only. The resulting power signal is hence the average real power over each 15-min time interval. To be consistent for the disaggregation, the power signals of the appliances were aggregated according to the categories presented above to create the reference ground truth. The reference whole-house power measurement is defined as the sum of the sub-measurements in order to ensure energy conservation. This implies that the noise is not considered due to the fact that averaging of the power measurement on a 15-min time window should minimize its influence. Additionally, as DUE algorithm disaggregates the whole house energy consumption, the noise should be included in one of the categories, most likely, the standby. Moreover, the metadata of the datasets were used to complete the household information requirements as briefly described in table \ref{tab:hh_ch}. The detail of the households information can be found in the appendix \ref{APP:HsldInfo}, table \ref{tab:HsldInfo}; the appliance list is described in table \ref{tab:HsldApp}. \par

Three publicly available datasets were selected: \textsc{eco}\cite{Beckel2014}, \textsc{smartenergy.kom}\cite{Alhamoud2014} and \textsc{uk-dale}\cite{Kelly}. The following section will briefly present these datasets.

\paragraph{The \textsc{ECO} dataset} 
\cite{Beckel2014,Beckel2014a}, made by ETH Zurich, was collected during a period of eight months (from the beginning of July 2012 to the end of January 2013) from six households in Switzerland. Registered measurements represent a new level of detail, comprising the voltage, current and phase shift between voltage and current readings for each of the three phases. This makes the dataset useful for algorithms that require both active and reactive power. The sampling rate of 1 Hz distinguishes this dataset from others in the field together with the information provided about the occupancy of the households. For this study, we used House n°2 from June \nth{1} to October \nth{30}, 2012.

\paragraph{\textsc{SMARTENERGY.KOM} dataset}
 \cite{Alhamoud2015} was developed to propose energy-savings recommendations based on detecting the activities performed by the user. The collection of energy consumption data by appliance is complemented by the measurement of motion (i.e. occupancy), temperature and brightness in the environment. The dataset accounts for more than 42 million data points for two households and it is the first dataset to combine power and environmental sensors' measurements with user feedback \cite{Alhamoud2014}.\\
The data from Apartment n°1 was collected for 82 days, while Apartment n° 2 participated in the experiment for 60 days. For both of the deployments, there are nine respective activities that should be recognized such as sleeping, watching TV, eating, ironing, reading, etc.
For the following, we used the Apartment n°1 from March \nth{5} to June \nth{25}, 2013.

\paragraph{\textsc{UK-DALE},}
 an open-access dataset for NILM research, is the first out of the UK with a high temporal resolution. To the best of our knowledge, this public dataset covers the longest period in Europe, from December 2012  to April 2017 for the latest release \cite{Kelly2017}. \\
The dataset records the active power demand from appliances as well as the whole-house apparent power for the five households that participated in the data collection. The sampling rate for both the main power and for the power of individual appliances is six seconds. Distinct from other public datasets for load disaggregation, \textsc{uk-dale} contains metadata as well. These additional data files include information such as the type of ownership, the number of inhabitants, inhabitants' characteristics, the heating type and any energy improvements made to the house. This information was fed into the proposed algorithm to complete the households' characteristics. \textsc{uk-dale} is included in the NILMTK \cite{Kelly2017,Batra2014a} framework. For the purpose of testing, we used the House n°1 during the period from April 2014 to April 2015. We chose this house because it was the most documented one and because a large share of appliances' consumption data is available. 

\subsubsection{Metrics}
\label{subsec:metrics}
In order to evaluate the performance of the proposed algorithm, it is necessary to define appropriate metrics that are consistent with the ultimate goal of this work. Typical performance metrics are based on event detection and come from classification algorithms literature \cite{Faustine2017}. The accuracy is defined as the ability of the algorithm to detect whether an appliance is on or off (see equation \ref{eq:Acc}).

\begin{equation}
\label{eq:Acc}
\textsc{Acc} = \frac{\text{Correct matches}}{\text{Total possible matches}} \\
\end{equation}

This metric is not appropriate, however, for appliances that are in one state  most of the time (like a TV is mostly off). In order to correct this, researchers often use the F measure \cite{Faustine2017} defined in equation \ref{eq:Fmeas}.

\begin{gather}
\text{Precision} =\frac{\text{TP}}{\text{TP}+\text{FP}} \\
\text{Recall} =\frac{\text{TP}}{\text{TP}+\text{FN}}\\
\label{eq:Fmeas}
F=\frac{2 \times \text{Precision} \times \text{Recall}}{\text{Precision}+\text{Recall}}
\end{gather}

\noindent where $\text{TP}$ = True Positive, $\text{FP}$ = False Positive and $\text{FN}$ = False Negative. \\

Recent attempts to have common performance metrics are based on similar metrics. Beckel \cite{Beckel2014} used the root mean square error of the $m^{th}$ appliance signal as defined in equation \ref{eq:RMSE}.

\begin{equation}
\label{eq:RMSE}
\textsc{RMSE}_m=\sqrt{\frac{1}{T}\sum_t{\left(\hat{P}_m^t-P_m^t\right)}^2}
\end{equation}

With this last metric the comparison between appliances having a high difference in power consumption (i.e a stove and an internet box) is problematic. For this reason, a derivation of the root mean squared error called the \textit{Normalized Disaggregation Error} ($\textsc{NDE}_m$)  \cite{Liu2018,Kolter2012,Dong2013} defined in equation \ref{eq:NDE},  normalizes the squared error of a single appliance by the total energy of the signal. Very similarly, Parson \cite{Parson2012} and Beckel \cite{Beckel2014} used a \textit{Normalized Error in the total Energy Assigned} ($\textsc{NEEA}_m$) as defined in equation \ref{eq:NDE2}.

\begin{align}
	\label{eq:NDE}
	\textit{Normalized Disaggregation Error: } & \textsc{NDE}_m = \frac{\sum_t {\left(\hat{P}_m^t - P_m^t \right)}^2}{\sum_t {\left(P_m^t\right)}^2}\\
	\label{eq:NDE2}
	\textit{Normalized Error in the total Energy Assigned: } & \textsc{NEEA}_m = \frac{\sum_t {|\hat{P}_m^t - P_m^t |}}{\sum_t {P_m^t}}
\end{align}

Makonin et al. in \cite{Makonin2015} used an \textit{Estimation Accuracy} ($\textsc{EstAcc}$), derived from \cite{Kolter} and \cite{Johnson2013},  of the appliance power signal or across all appliances as in equation \ref{eq:AccM} and \ref{eq:AccAll} respectively. 

\begin{gather}
\label{eq:AccM}
\textsc{EstAcc}_m=1-\frac{\sum_t|\hat{P}_m^t-P_m^t|}{2\cdot\sum_t P_m^t}\\
\label{eq:AccAll}
\textsc{EstAcc}=1-\frac{\sum_t\sum_m|\hat{P}_m^t-P_m^t|}{2\cdot\sum_t\sum_m P_m^t}
\end{gather}

In this paper, we choose the \textit{Estimation Accuracy} as a reference metric to compare the disaggregation performance, since it is a popular metric in recent literature. The use of the root mean squared error would not be appropriate because it is not possible to use it for multiple appliance comparison for the reason explained above. Finally, the original benchmark performed with the normalized disaggregation error has shown similar trend as what will be discussed in the following section. So this metric was given away for clarity reason.  In addition, because the aim of our methodology is to infer the energy share by category of appliances, the \textit{Energy Share Error} per category is defined in equation \ref{eq:ErrCatSh}.

\begin{equation}
\label{eq:ErrCatSh}
 \textit{Energy Share Error: }   \textsc{ESE}_m =\frac{\sum_t\hat{P}_{m}^t}{\sum_t \sum_{m} \hat{P}_{m}^t} -\frac{\sum_t P_{m}^t}{\sum_t \sum_{m} P_{m}^t}
\end{equation}

\noindent Note that in the following, the index $m$ corresponds to an appliance category as defined in table \ref{tab:AppData}.\\

\section{Results and discussion}
\label{sec:Results}

We compared the performance of our proposed method against the four common algorithms of the NILM literature presented above. 
Testing was performed using three datasets in order to evaluate the spread in performance of all algorithms. As described in section \ref{subsec:metrics}, relevant metrics in the context of flexibility estimation are the \textit{Energy Share Error} per category and the \textit{Estimation Accuracy}. Additionally, the execution times for training and testing were also compared. \\
Each test was conducted by dividing the dataset into a training period and a testing period. The length of the training period is approximately two times the length of the testing period. As our proposed algorithm does not require any training, it was tested on the testing period only. The duration of the training and testing periods for different datasets is reported in the following table: 

\begin{table}[h!]
	\centering
    \small
	\caption{\label{tab:TrainTestPeriod} \bf Testing and training periods for the three datasets}
\begin{tabular}{@{}llllllll@{}}
	\toprule
	& \multicolumn{3}{c}{\textbf{Training period}} && \multicolumn{3}{c}{\textbf{Testing period}} \\ 
    \cmidrule{2-4} \cmidrule{6-8}
    & from & to & days && from & to & days\\ 
	\midrule
	\textbf{\textsc{ECO}} & 2012-06-01 & 2012-09-30 & 121 && 2012-10-01 & 2012-10-30 & 29\\ 
	\textbf{\textsc{SMARTENERGY.KOM}} & 2013-03-05 & 2013-05-19 & 75 && 2013-05-19 & 2013-06-25 & 37\\ 
	\textbf{\textsc{UK-DALE}} & 2014-04-01 & 2015-04-01 & 365  && 2015-04-01 & 2015-07-01 &91 \\ 
	\bottomrule	
\end{tabular} 
\end{table}

\subsection{Performance metrics}
The two performance metrics chosen to compare the algorithms are the \textit{Energy Share Error} and the \textit{Estimation Accuracy}. These metrics are reported in this section and analyzed with respect to their relevance. \par
The energy share by category is defined as in equation \ref{eq:EcatSh}:

\begin{equation}
\label{eq:EcatSh}
\hat{S}_m=\frac{\sum_t\hat{P}_{m}^t}{\sum_t \sum_{m} \hat{P}_{m}^t}
\end{equation}

\begin{figure}[H]
\centering
\subfloat[\textsc{ECO} dataset]{\label{fig:ECO_E} \includegraphics[width=0.9\textwidth]{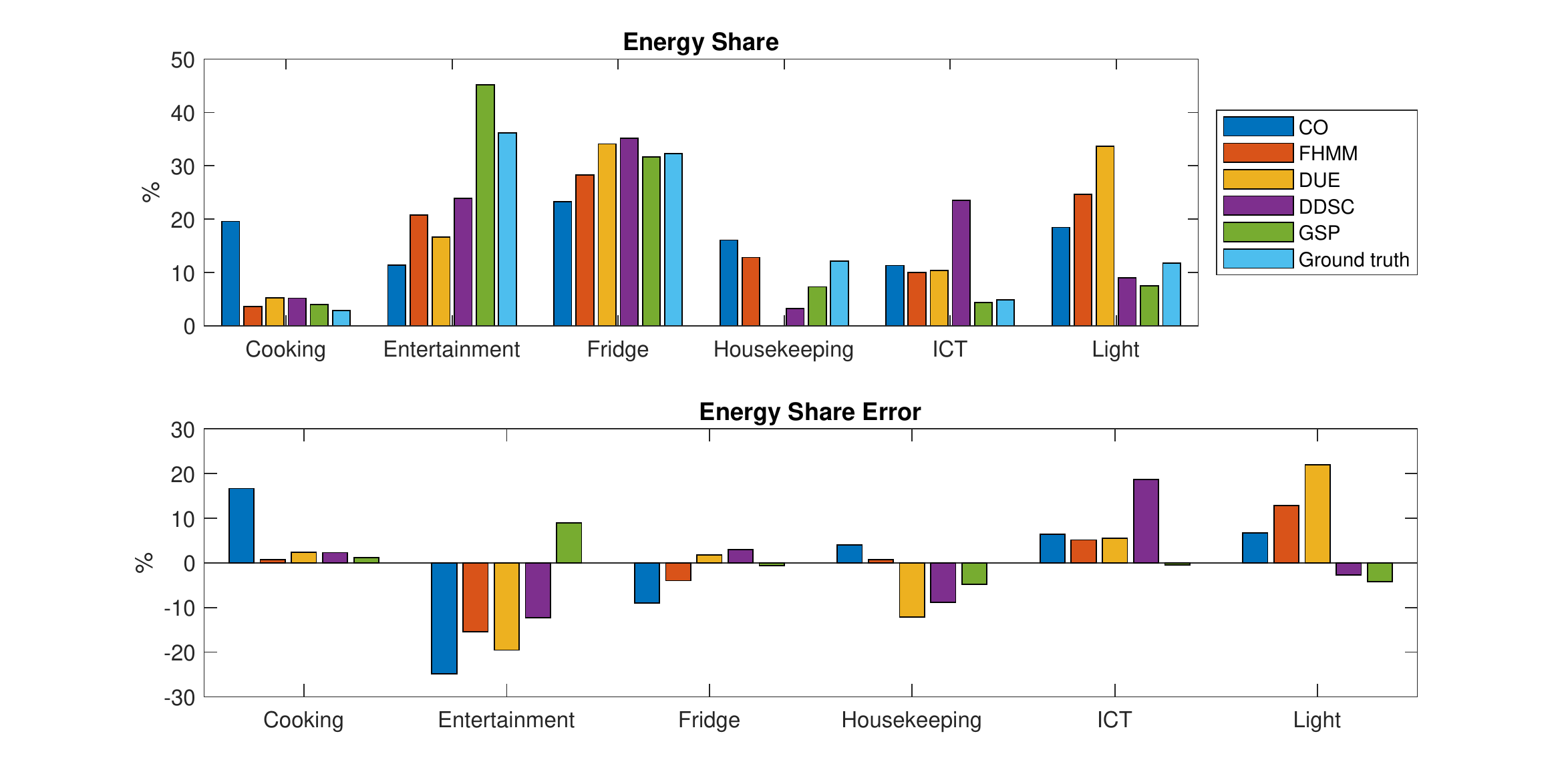}}
\hspace{5pt}
\subfloat[\textsc{SMARTENERGY.KOM} dataset]{\label{fig:SME_E} \includegraphics[width=.9\textwidth]{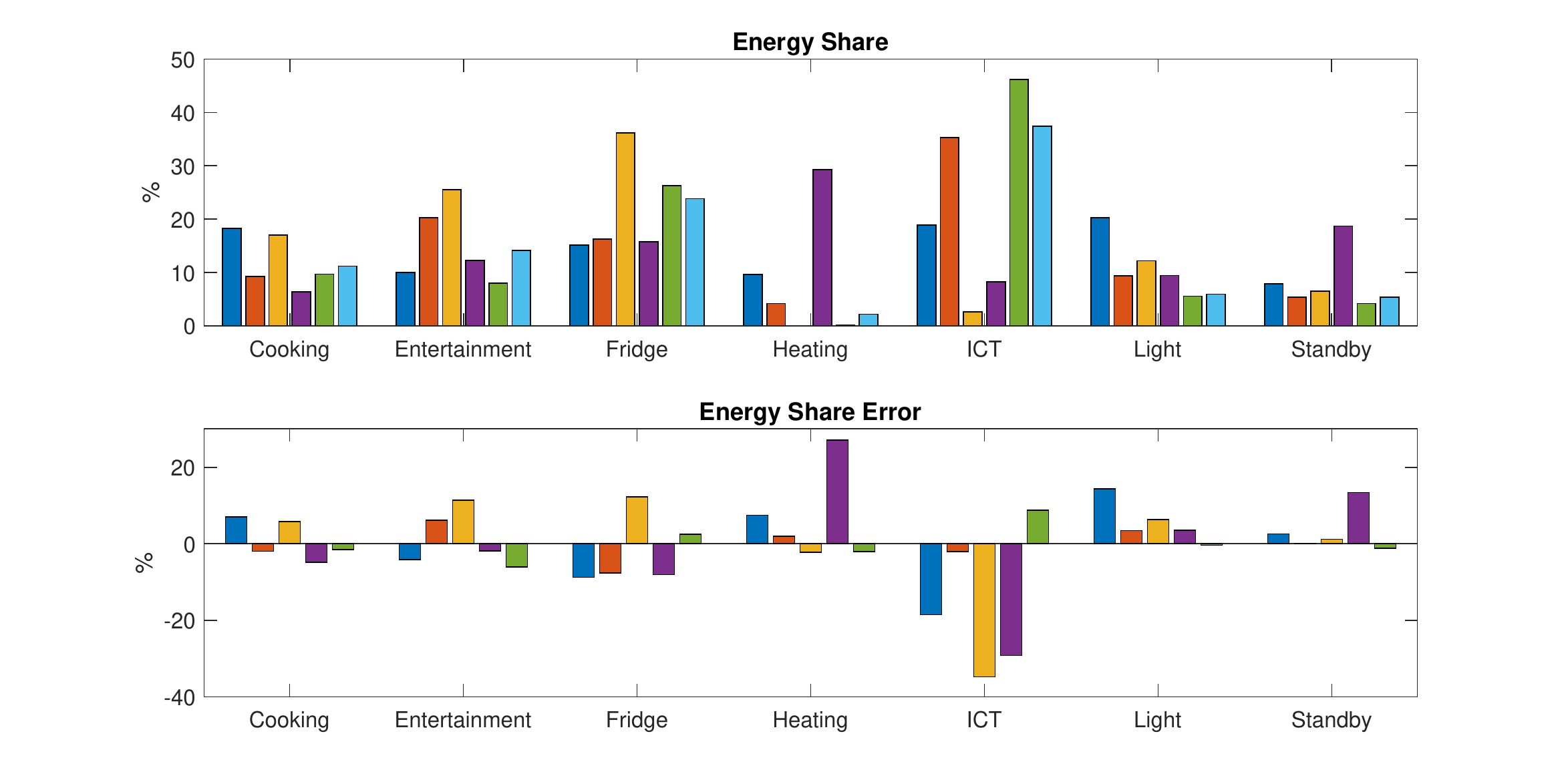}}
\hspace{5pt}
\subfloat[\textsc{uk-dale} dataset]{\label{fig:UKDALE_E} \includegraphics[width=.9\textwidth]{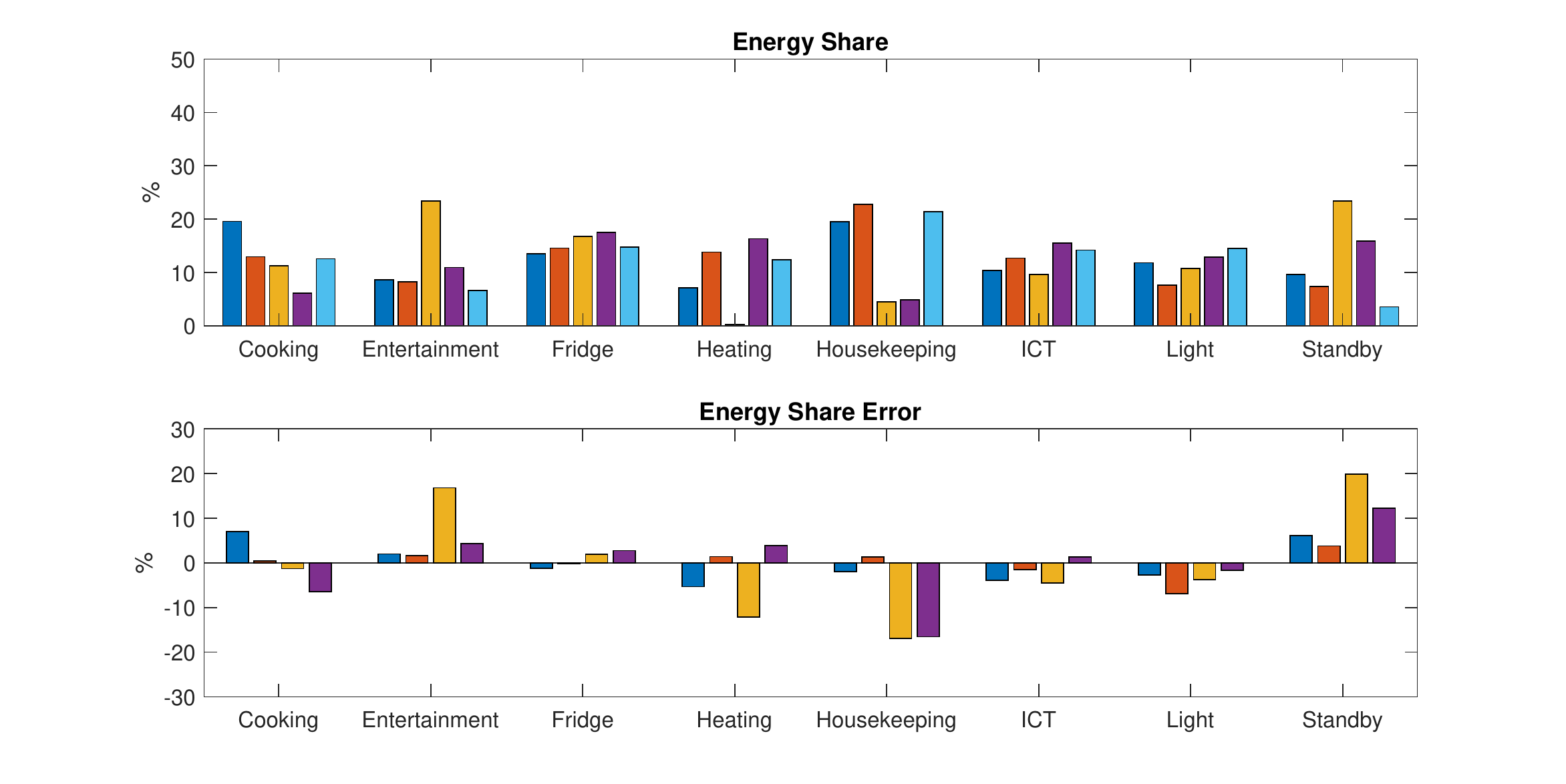}}
\caption{\label{fig:E_share} \bf Energy share and error.}
\end{figure}

The \textit{Energy Share Error} for each category, as defined in equation \ref{eq:ErrCatSh}, are reported in table \ref{tab:ErrCatSh}. ''NA'' means that this category was not measured in the respective dataset. In the GSP columns, all values corresponding to the \textsc{uk-dale} dataset are NAs because this algorithm was not run on this dataset. Indeed, the complexity of the GSP algorithm increases non-linearly with the length of training and testing periods, therefore it is more appropriate for small training periods as shown in \cite{He2016}. Although various strategies have been tested,  we have not been able to find a satisfying implementation of the GSP algorithm that would allow performing disaggregation on a large dataset. A comparison of the performance for each dataset with respect to this metric is shown in figure \ref{fig:E_share}. A positive error means an overestimation of the energy share. The \textit{Energy Share Error} is highly dependent on both the dataset and the observed categories. Surprisingly, the \textit{Energy Share Error} is especially high for the ICT category in the \textsc{smartenergy.kom} dataset. This may be due to the unusually high share of energy consumed by this category (about 40\%, whereas it is below 20 \% in the other two datasets). The estimation of light consumption is too high for the DUE algorithm for the \textsc{eco}  and \textsc{smartenergy.kom} datasets. One has to consider that the datasets are often incomplete, i.e., they do not represent the whole and real power consumption of a house. Authors of such datasets actually choose the appliances to be monitored depending on their capabilities while neglecting the appliances that are too difficult or impossible to monitor. This is also the reason why no Housekeeping is present in the \textsc{smartenergy.kom} dataset. As the DUE algorithm is based on surveys and statistics, it always assumes that all categories are present in the house, thus explaining these differences. \par

\begin{table}[h!]
	\centering
    \small
	\caption{\label{tab:ErrCatSh} \bf \textit{Energy Share Error} per category (in $\%$) {\footnotesize (\textsc{ECO}, \textsc{SMARTENERGY.KOM}, UK-DALE)}}
\begin{tabular}{@{}lccccc@{}}
	\toprule
    & \textbf{CO} &\textbf{FHMM} & \textbf{DUE} & \textbf{DDSC} & \textbf{GSP} \\
	\midrule
	\textbf{Cooking} & ( 17, 7, 7)  & (1 , -2, 0)  & ( 2 , 6, -1)  & ( 2, -5, -6)  & ( 1, -2 , NA) \\
	\textbf{Entertainment} & (-25, -4, 2)  & (-15 , 6, 2)  & (-20 , 11, 17)  & (-12, -2, 4)  & ( 9, -6, NA) \\
    \textbf{Fridge} & ( -9, -9, -1)  & ( -4, -8, 0)  & ( 2, 12, 2)  & ( 3, -8, 3)  & ( -1, 2, NA) \\
    \textbf{Heating} & ( NA, 7, -5)  & ( NA, 2, 1)  & ( NA, -2,-12)  & ( NA, 27, 4)  & ( NA, -2, NA) \\
    \textbf{Housekeeping} & ( 4, NA, -2)  & ( 1, NA, 1)  & (-12, NA,-17)  & ( -9, NA,-17)  & ( -5, NA, NA) \\
    \textbf{ICT} & ( 6,-19, -4)  & ( 5, -2, -2)  & ( 6,-35, -5)  & ( 19,-29, 1)  & ( -4, 0, NA) \\
    \textbf{Light} & ( 7, 14, -3)  & ( 13, 3, -7)  & ( 22, 6, -4)  & ( -3, 4, -2)  & ( -4, 0, NA) \\
    \textbf{Standby} & ( NA, 3, 6)  & ( NA, 0, 40)  & ( NA, 1, 20)  & ( NA, 13, 12)  & ( NA, -1, NA) \\
    \bottomrule	
\end{tabular} 
\end{table}

\begin{figure}[H]
\centering
\includegraphics[width=1\textwidth]{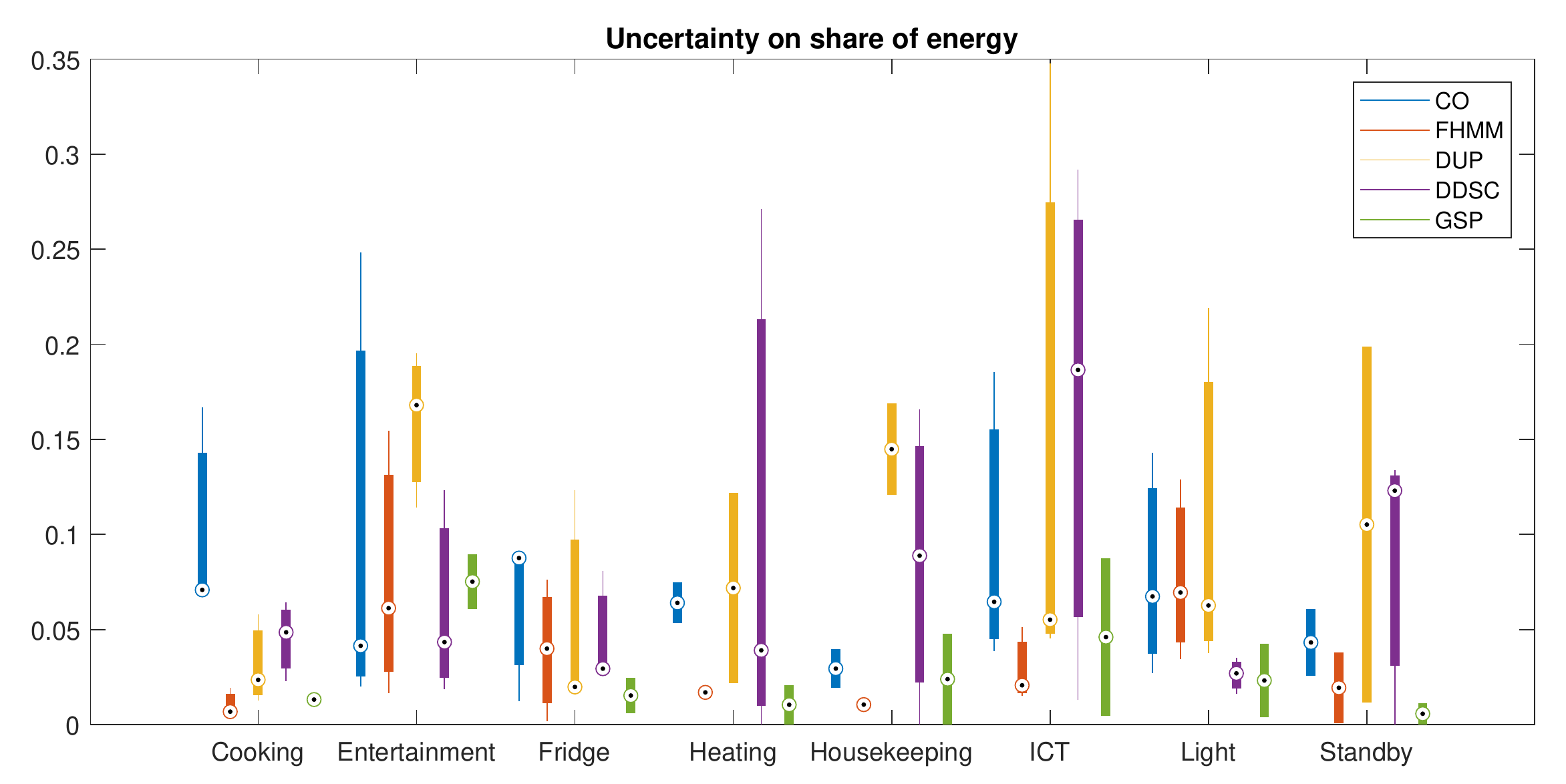}
\caption{\label{fig:All_E_Uncert} \bf Energy share uncertainties across all dataset.}
\end{figure}

The global uncertainty (here defined as the error on the energy share) is in the range of 20\% for all algorithms except the DUE as depicted in figure \ref{fig:All_E_Uncert}. The uncertainty is lower on average for the supervised algorithms, especially for the FHMM. The good performance of the GSP algorithm has to be counter-balanced by the fact that this algorithm has not been tested on the \textsc{uk-dale} dataset. Taking into account that not all categories were present in each of the two other datasets, only a few data points were extracted to present these metrics. 

The \textit{Energy Share Error} per category is a global metric that does not take into account the temporal accuracy. To focus on this aspect, the \textit{Estimation Accuracy} is more appropriate. The \textit{Normalized Disaggregation Error} could serve as an equivalent metric, however, the \textit{Estimation Accuracy} metric is more commonly seen in the recent literature, explaining its usage in the present benchmark. Moreover, the following outcomes would have been similar if the \textit{Normalized Disaggregation Error} had been used.

\begin{figure}[H]
\centering
\subfloat[\textsc{ECO} dataset]{\label{fig:ECO_EstAcc} \includegraphics[width=0.8\textwidth]{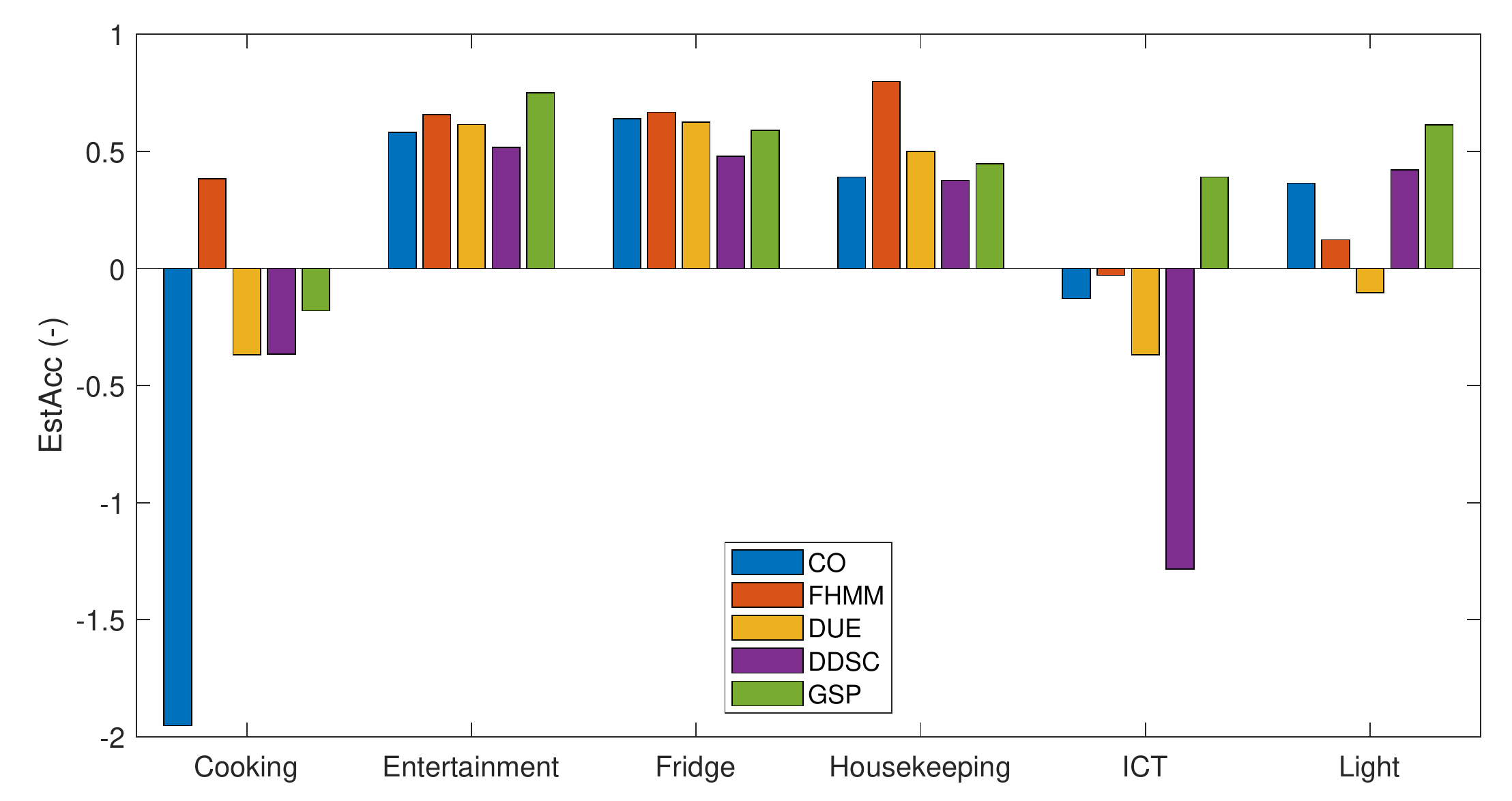}}
\hspace{5pt}
\subfloat[\textsc{SMARTENERGY.KOM} dataset]{\label{fig:SME_EstAcc} \includegraphics[width=.8\textwidth]{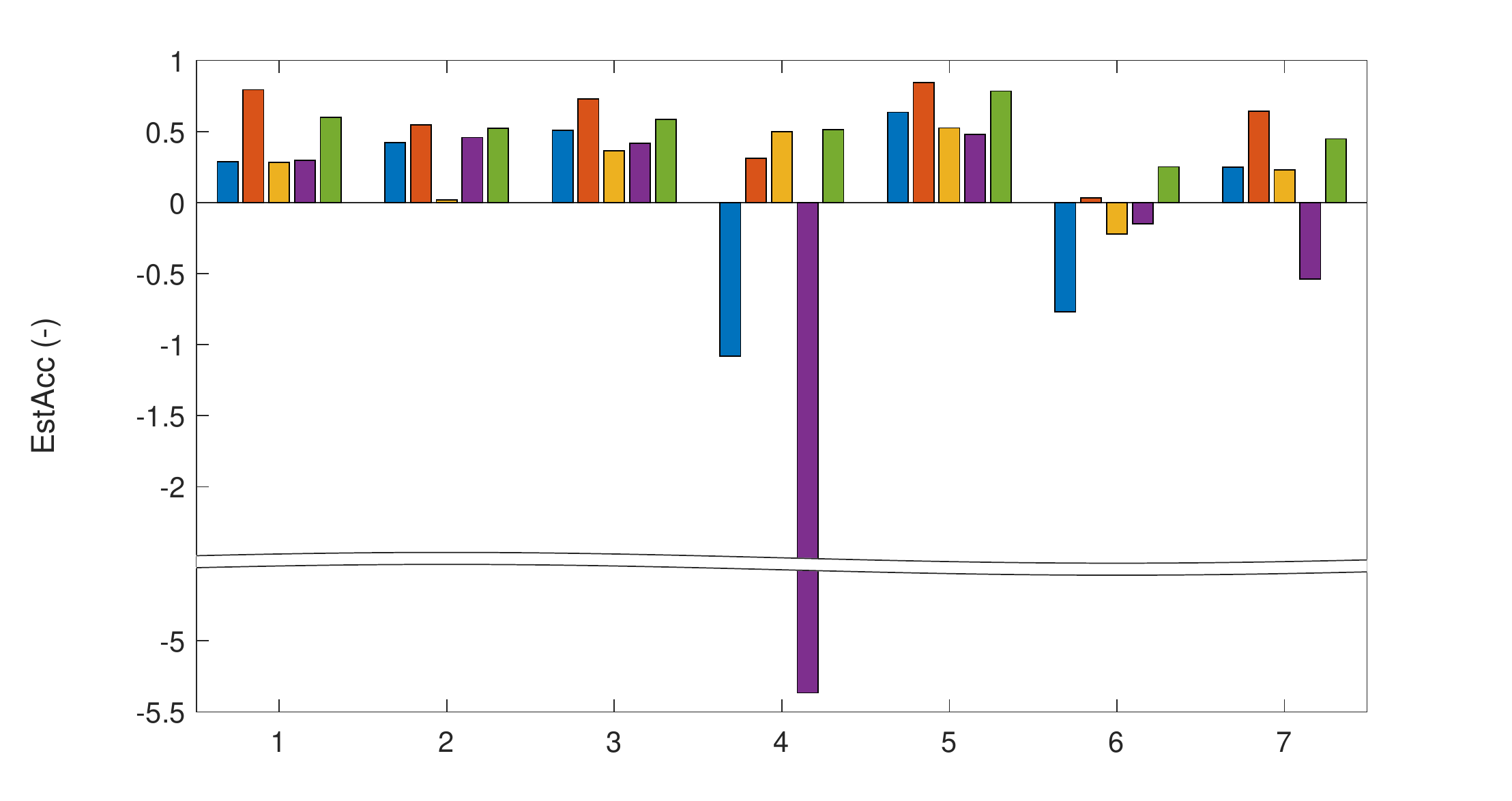}}
\hspace{5pt}
\subfloat[\textsc{uk-dale} dataset]{\label{fig:UKDALE_EstAcc} \includegraphics[width=.8\textwidth]{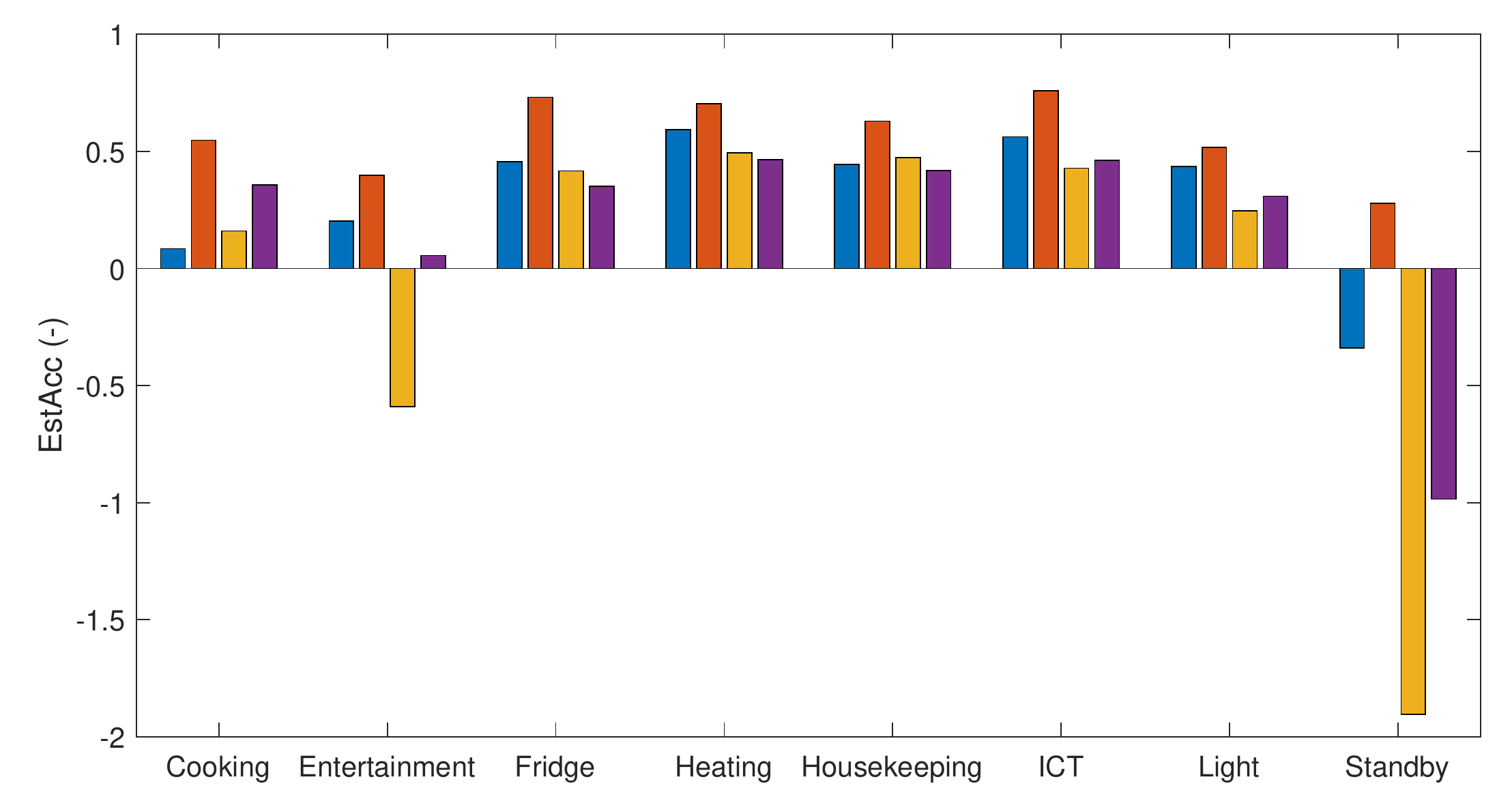}}
\caption{\label{fig:EstAcc} \bf \textit{Estimation Accuracy}.}
\end{figure}

The \textit{Estimation Accuracy} for each category are reported in table \ref{tab:EstAcc}. Figure \ref{fig:EstAcc} shows this metric per category for each dataset and algorithm. Negative values indicate a poor disaggregation result, occuring when the sum of the absolute errors is greater than two times the energy of the signal. The y-axis has been intentionally cropped between -2 and 1 for the sake of comparison across datasets. All algorithms experienced negative values for this metric but not always on the same categories. Looking specifically at categories that seem easier to disaggregate (all \textit{Estimation Accuracy} are greater than zero), the DUE algorithm performs averagely as well as the other algorithms. Taking into account that the DUE algorithm relies only on statistical information to infer the power signal of each category, the temporal accuracy, especially for categories linked to short-duration activities like ICT and Entertainment are difficult to catch in unsupervised way. Some extreme negative values are difficult to explain (for instance, the disaggregation of cooking in the \textsc{eco} by the CO algorithm). As regards with the Standby category, the poor performance of the DUE algorithm can be explained by the fact that DUE considers standby as a constant load, while in the ground truth it is not constant and considered as any other appliance by the other algorithms.   

\begin{table}[h!]
	\centering
    \small
	\caption{\label{tab:EstAcc} \bf \textit{Estimation Accuracy} per category (\%)  {\footnotesize (\textsc{ECO}, \textsc{SMARTENERGY.KOM}, \textsc{UK-DALE})}}
\begin{tabular}{@{}lccccc@{}}
	\toprule
    & \textbf{CO} &\textbf{FHMM} & \textbf{DUE} & \textbf{DDSC} & \textbf{GSP} \\
	\midrule
	  
    \textbf{Cooking} & (-195, 29, 8) & (38, 79, 55) & (-37, 28, 16) & (-37, 30, 36) & (-18, 60, NA) \\ 
    \textbf{Entertainment} & (58, 42, 20) & (66, 55, 40) & (62, 2, -59) & (52, 46, 6) & (75, 52, NA) \\ 
    \textbf{Fridge} & (64, 51, 46) & (67, 73, 73) & (62, 36, 42) & (48, 42, 35) & (59, 58, NA) \\ 
    \textbf{Heating} & (NA, -108, 59) & (NA, 31, 70) & (NA, 50, 49) & (NA, -537, 46) & (NA, 51, NA) \\ 
    \textbf{Housekeeping} & (39, NA, 44) & (80, NA, 63) & (50, NA, 47) & (38, NA, 42) & (45, NA, NA) \\ 
    \textbf{ICT} & (-13, 63, 56) & (-3, 85, 76) & (-37, 52, 43) & (-128, 48, 46) & (39, 78, NA) \\ 
    \textbf{Light} & (36, -77, 44) & (12, 3, 52) & (-10, -22, 25) & (42, -15, 31) & (61, 25, NA) \\ 
    \textbf{Standby} & (NA, 25, -34) & (NA, 64, 28) & (NA, 23, -190) & (NA, -54, -98) & (NA, 45, NA) \\ 
    
    \bottomrule	
\end{tabular} 
\end{table}

One has to remember that forecasting a constant zero power signal leads to an \textit{Estimation Accuracy} of 0.5. This is typically achieved by the DUE algorithm for the Heating category on the \textsc{smartenergy.kom}. This is the drawback of this metric. It is easy to interpret a value close to one as a good performance, but it is hard to interpret the quality of an algorithm that produces an \textit{Estimation Accuracy} close to zero or even negative. 

To summarize, careful analysis of the \textit{Energy Share Error} and the \textit{Estimation Accuracy} shows that the DUE algorithm performs similarly to other algorithms although its temporal accuracy suffers from the statistical approach. As the goal of this algorithm is to be used by a utility on a large number of households, it must have an acceptable computational cost in addition to a reasonable accuracy. The following section aims at comparing the algorithms taking into into account these considerations. 

\subsection{Execution time }
The execution time is reported in table \ref{tab:ExecTime}. As expected, execution time scales with the length of the dataset (in terms of both training and testing periods). The CO algorithm, which has the lowest complexity, is always the fastest to execute, as also reported by  Manivannan et al.\cite{reviewer2}. One should keep in mind that the GSP algorithm was not tested on the \textsc{uk-dale} dataset due to the algorithm's characteristics. Namely, GSP performs training and disaggregation in a single step (for each category of appliances) by constructing a graph and solving the optimization problem (as described in section \ref{subsubsec:algo}). This step requires a lot of computational effort, hence the execution time becomes too long compared to other algorithms. For short datasets, the DUE algorithm has an execution time comparable to those from FHMM and DDSC. However, these algorithms scale badly with the length of the dataset (or require advanced parallelization techniques, which were not implemented here). Due to its sequential nature, the execution time of the DUE algorithm increases linearly with the length of the dataset.

\begin{table}[H]
\centering
\caption{\label{tab:ExecTime} \bf Execution time.}
\begin{tabular}{@{}lcccccccc@{}}
\toprule
     &  \multicolumn{2}{c}{\bf \textsc{ECO}} && \multicolumn{2}{c}{\bf \textsc{SMARTENERGY.KOM}} && \multicolumn{2}{c}{\textbf{\textsc{UK-DALE}}} \\
 \cmidrule{2-3} \cmidrule{5-6} \cmidrule{8-9}    
& training			& testing	&& training		& testing	&& training		& testing\\
& 121 days	& 29 days && 75 days	& 37 days && 365 days	& 91 days\\
\midrule
\bf CO   & 2s       & $<$1s 	&& $<$1s      	& $<$1s  	&& $<$1s      	& 9s\\
\bf FHMM & 17s      & 7s       	&& 13s          & 88s   	&& 58s          & 1h \\
\bf DUE  &          & 59s 		&&              & 45s     	&&	  			& 6min\\
\bf DDSC & 42s   	& $<$1s		&& 19s  		& $<$1s		&& 5min			& 6s \\
\bf GSP  & \multicolumn{2}{c}{9h} && \multicolumn{2}{c}{4h}  && \multicolumn{2}{c}{NA}\\
\bottomrule
\end{tabular}
\end{table}

\begin{figure}[H]
	\centering
	\includegraphics[width=0.8\textwidth]{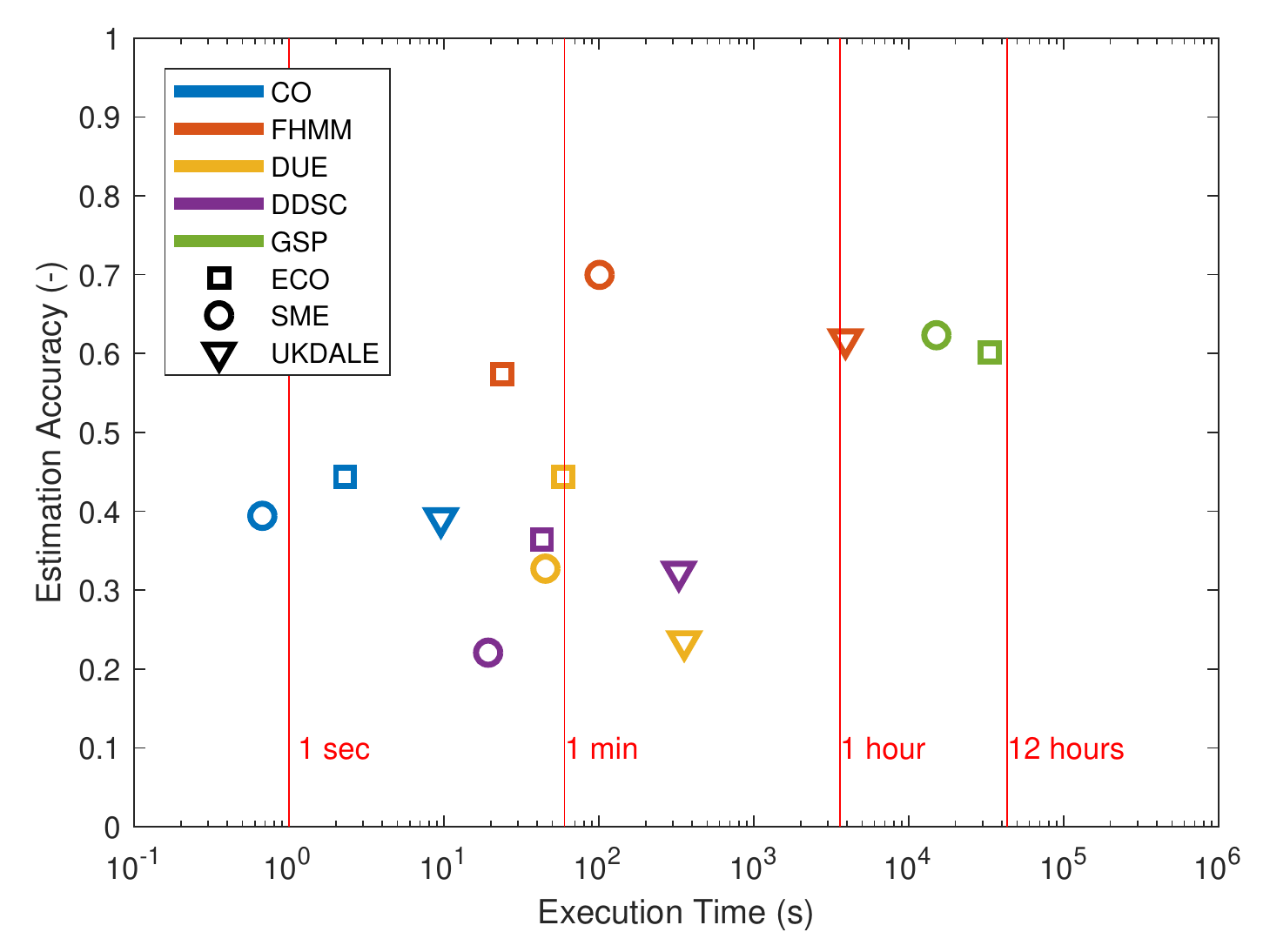}
	\caption{\label{fig:EstAccVSTime} \bf \textit{Estimation Accuracy} versus execution time.}
\end{figure}

Figure \ref{fig:EstAccVSTime} depicts the global \textit{Estimation Accuracy} (defined in equation \ref{eq:AccAll} from section \ref{sec:Meth}) as a function of the total execution time, i.e the sum of the testing time and training time if any. Clearly, FHMM is the best compromise as the gain in accuracy according to this metric is significant although the execution time is one order of magnitude greater than the DUE for the longest dataset (\textsc{uk-dale}). 

\subsection{Discussion}
These results confirm the fact that supervised algorithms perform better than unsupervised algorithms. Although the disaggregation uncertainty is generally higher for the DUE than for the best supervised algorithm, it falls within the same range. The \textit{Estimation Accuracy} is used to assess both the magnitude and temporal accuracy of the algorithm on the dataset and shows that DUE performs averagely with respect to other algorithms. For large scale disaggregation of households' energy consumption, it is not always possible to monitor all households at the appliance level as it was done in \cite{Hakell2015}. Similarly, there is not always a set of reference houses with power consumption data at the appliance level (as in \cite{Batra2016}), on which training can be achieved before performing disaggregation of whole-house power consumption. In the case where the only available information is the households' characteristics and smart meter measurements, the DUE algorithm is able to disaggregate the energy consumption with an uncertainty range comparable to supervised NILM algorithms and does not suffer from computational limitations. 

Future investigations should focus on the sensitivity of the DUE algorithm with respect to household characteristics. A probable source of uncertainty is that some households' information is not known precisely from the dataset metadata. Therefore, this information was deduced, assumed or extrapolated. As the algorithm relies on a statistical approach, it does not require exact information to perform disaggregation. However, no dedicated study has yet been conducted to determine to which extend this information has to be exact and reliable in order to extract the best performance. Indeed, it appears from the literature that very few studies include non-electrical based input features to increase disaggregation accuracy (see table \ref{tab:hh_ch}). Only \cite{Batra2016} included households data to aid the disaggregation process. Future research should investigate the relationship between disaggregation accuracy, variety of the input features and sampling rates. The proposed input features of the algorithm are very common to most social studies and could be easily acquired in any context. However, if more features are added to the model such as exact occupants' timetables, the question of data accessibility and privacy will arise.  

As the whole-house power signal is assumed to be the sum of each category's sub-signal, the resulting synthetic power measurement does not correspond exactly to the real whole-house power measurement. Indeed, each dataset covers a certain number of individual appliances, which were grouped to form the categories. For \textsc{uk-dale}, about 20 appliances were monitored, while for two other datasets this number is approximate to 10. In all cases, this is not representative of a full coverage of the electric power consumption of a house. As a result, this might lead to some error in the DUE. By design, it assumes that the input power measurement, as the whole-house power consumption, is representative of the full household energy activities. Hence providing partial electrical power information of the household activities might have a significant impact on the performance of the algorithm. However, it is very difficult to evaluate this claim, as to our knowledge, currently no dataset has a full coverage of all appliances used by a household over a period of several weeks.

\section{Conclusion}
\label{sec:Conclusion}
A new hybrid method for non-intrusive appliance load monitoring is proposed. The \textit{device usage estimation} method is based on the generation of an activity chain using a Markov model while adapting the transition probability by restricting possible activities according to the power budget for the next time-step. The approach is hybrid by its nature, being a mix between a load simulation tool and a disaggregation algorithm. This feature makes it possible to disaggregate whole-house power measurement at a very low sampling rate. The proposed algorithm was tested using three different datasets down-sampled at a 15-min time resolution and benchmarked against four state-of-the-art non-intrusive load monitoring algorithms: the factorial hidden Markov model, combinatorial optimization, discriminative disaggregation via sparse coding, and graph signal processing. The results show better performance on average for the state-of-the-art algorithms as all of them are supervised and were trained on each dataset before being tested. Emphasizing that our \textit{device usage estimation} method is unsupervised, its prediction uncertainties normally remain under 20\%. The advantages of the proposed  algorithm include the absence of a training requirement and its low demand for computational power compared to the state-of-the-art non-intrusive load monitoring algorithms. This method is hence a novel unsupervised disaggregation method suitable for very low sampling rates. \par
A research application is to use this method for household demand-side management potential estimation. The approach would be to disaggregate the power measurements of a pool of households into categories, then classify the categories according to their demand-side management potential.  An industrial application could the use by a distribution system operator  to provide feedback and advice to its customers while saving the cost of installing additional devices. As most of the \textit{device usage estimation} input data are associated with the household typology, this information could be easily deduced from the building type itself, further simplifying the use of the DUE algorithm for utilities. The only requirement is an adapted smart-metering infrastructure. Future work should include consideration of adding more categories and optimizing the method in order to replace the sequential approach by an integrated optimization method.

\section* {Acknowledgments}
This work was supported by the Swiss Federal Office of Energy (SFOE) in the framework of the Flexi project, InnoSuisse in the framework of the SCCER-FURIES and The European Union's Horizon 2020 research and innovation program in the framework of the FEEdBACk project (grant agreement No 768935). We would also like to thank Alessandro Cuozzo for his contribution to the algorithm's description.
\clearpage
\section*{Appendix}
\appendix
\section{"Pre-treatment" procedure }
\label{APP:PreTreatment}
\begin{figure}[H]
\centering
\includegraphics[scale=\WFscale]{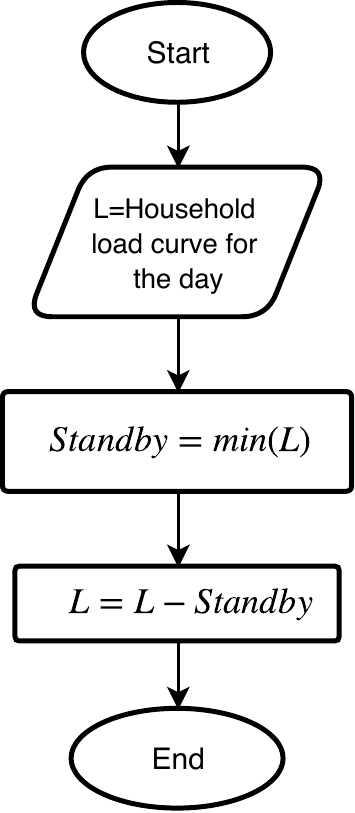}
\caption{\label{fig:WFstandby} \bf Standby filtering.}
\end{figure}

The pre-treatment procedure consists of filtering out the standby consumption and the fridge consumption (as shown in figure \ref{fig:WF_main}). Retrieving the standby power from the aggregated load curve is straightforward. The minimum value of the load for the current day is considered to be the standby consumption. This value is then subtracted from the aggregated power signal as shown in figure \ref{fig:WFstandby}. 

\begin{figure}[H]
\centering
\includegraphics[scale=\WFscale]{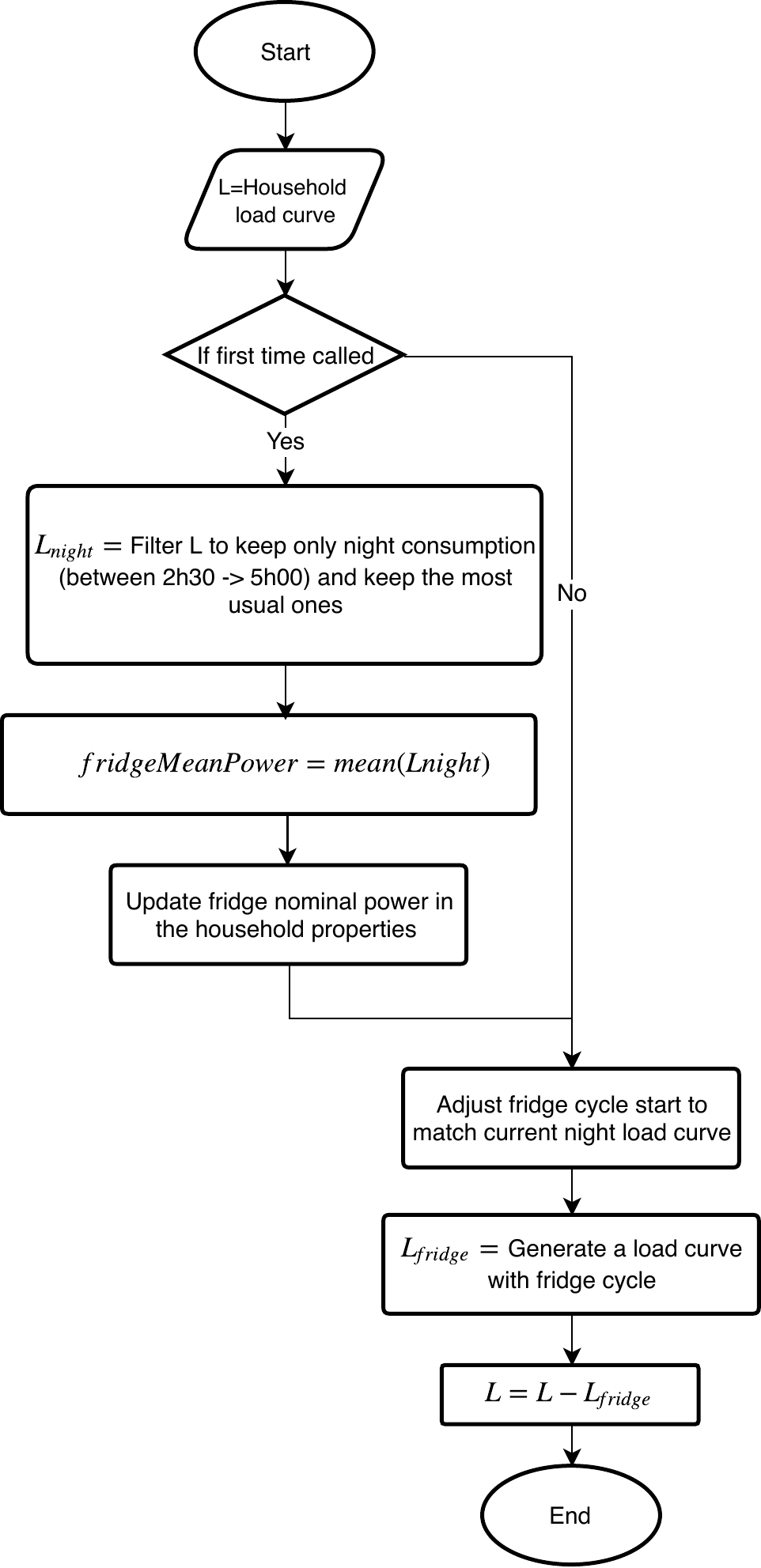}
\caption{\label{fig:WFfridge} \bf Fridge filtering flow chart.}
\end{figure}

Filtering out the fridge consumption pattern involves a few more steps. The main workflow is represented in figure \ref{fig:WFfridge}. On the first call (i.e. for the first considered day), a filter is applied to extract all nights of the dataset (not only the first night). Night starts at 02h30 and ends at 05h00. The resulting samples are further clustered based on the mean power consumption (using simply the Matlab \texttt{hist} command). The mean power of the largest cluster is then considered to be the mean power consumption of the fridge during the night. At the end of this procedure, the actual fridge nominal power is estimated and saved in the household inventory according to the following expression: 

\begin{equation}
P_\text{fridge New}=P_\text{fridge Old}\cdot\frac{\text{fridge mean power}}{P_\text{fridge Old}\cdot\beta_2}
\end{equation}

\noindent where $P_\text{fridge Old}$ is the default nominal fridge power and $\beta_2$ is the duty cycle of the fridge during the night. Both are reported in table \ref{tab:AppData}. \\

Finally the fridge cycle is adjusted by  minimizing, for the cycle start time and the cycle length, the sum of the squared error between the measured power signal at night ($L_\text{night}$) and the simulated power signal ($L_\text{fridge}$). 

\section{Random activity chain}
\label{APP:ActChain}
\begin{figure}
\centering
\includegraphics[scale=\WFscale]{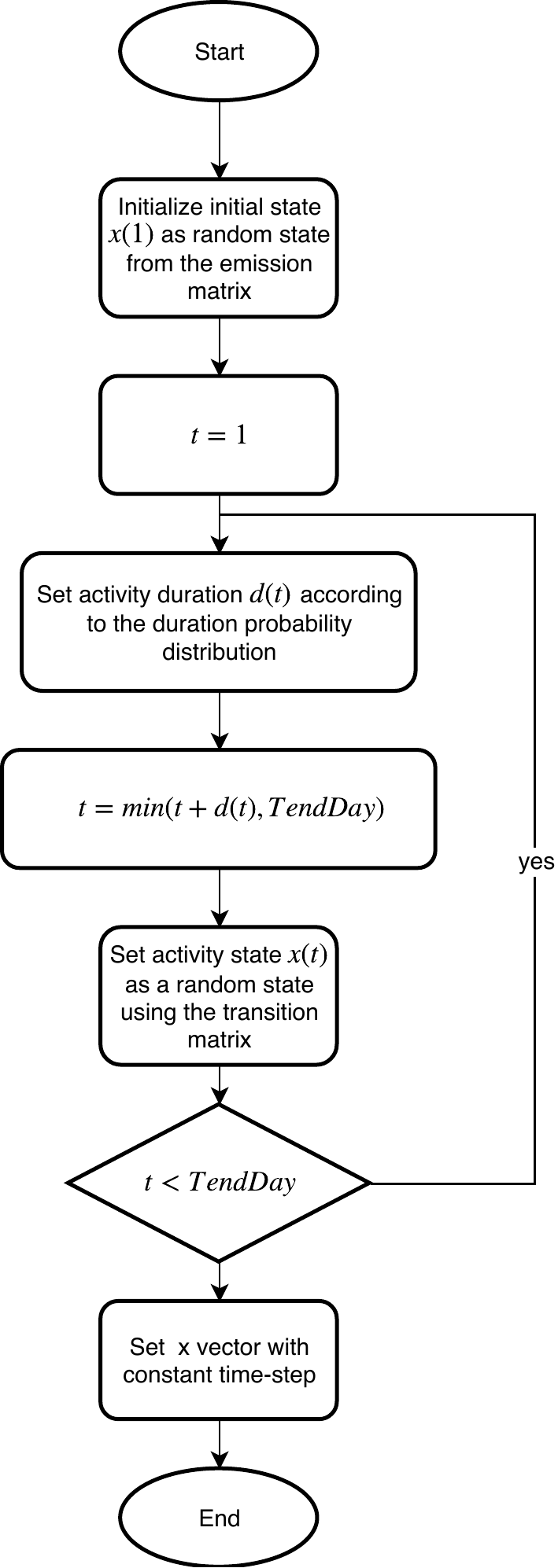}
\caption{\label{fig:WFgenRandAct} \bf Generation of random activity.}
\end{figure}

The generation of an activity chain is  performed by generating random integers where each integer value corresponds to the activity state $x^t \in S$. The probability distribution is given either by the transition matrix $A$ or by the initial probability distribution $\pi $. Similarly, the activation duration is generated using a discrete probability distribution. The sequence for generating the activity chain is depicted in figure \ref{fig:WFgenRandAct}. 

The algorithm to generate a random integer from a given discrete probability distribution $f(n),n=1...N$ (corresponding to one column of $A$) is the following: 

\begin{equation}
x_\text{rand}=\text{find first} \; n \text{ such that } \left(\epsilon \leq \frac{F(n)}{\sum_{i=1}^N f(i)}\right)
\end{equation}

\noindent with $F(n)$ being the cumulative probability distribution i.e. $F(n)=\sum_{i=1}^n f(i)$, and $ \epsilon $ a pseudo uniformly distributed random number ($\epsilon \in \left]0,1\right[$) from the Matlab \texttt{rand()} command.

\section{"Recognize load profile" step}
\label{APP:Recog}
This section aims at providing some details about the subprocess of the "Recognize load profile" step depicted in figure \ref{fig:WF_recog_main}.

At the start of the process, the aggregated load curve and activity chain for the considered person are the two main input arguments. Each subprocess also includes a device state vector that is not always explicitly mentioned for clarity. This state vector prevents the use of the same appliance by two people at the same time and hence an overestimation of the consumption of this category. Additionally  when several of the same appliances are in the house, the state vector is useful to take into account the fact that a similar appliance might be already in use when simulating a new device power signal. 

\paragraph{"Recognize Heating".}
The current version does not consider yet central electrical heating. Only a hairdryer is explicitly extracted. One reason is that the datasets of interest do not contain electrical heating systems. 

\begin{figure}[H]
\centering
\includegraphics[scale=\WFscale]{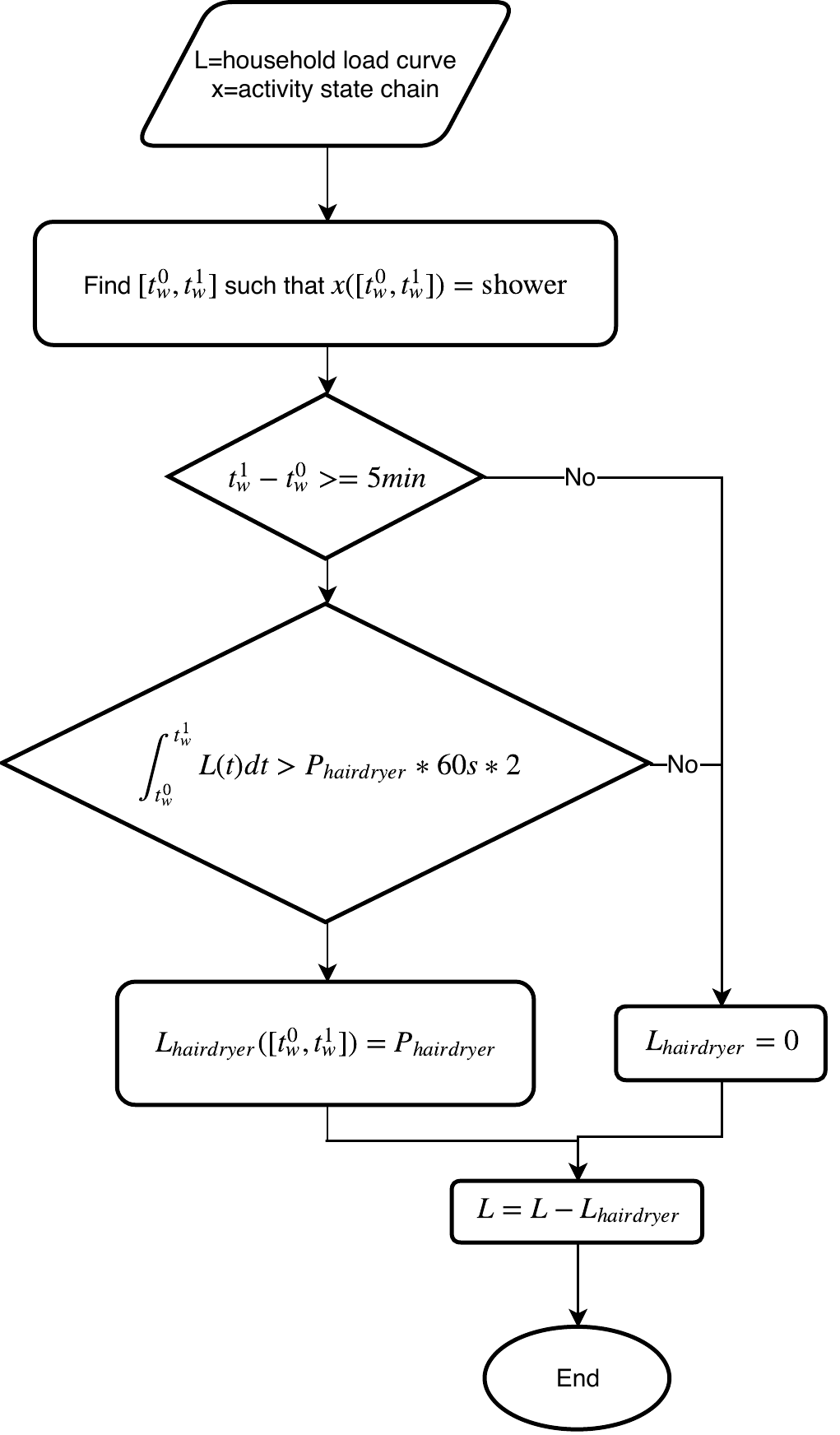}
\caption{\label{fig:WFrecogHeating} \bf Recognize Heating subprocess.}
\end{figure}

The first step consists of detecting a period $[t_w^0 t_w^1]$ where the activity corresponds to  \textit{Shower} , then check that the duration of this activity  is at least 5 min. The core of this process, depicted in figure \ref{fig:WFrecogHeating}, is to check that there is enough energy consumed between $t_w^0$ and $t_w^1$ for the hairdryer to run. Finally the corresponding power signal $L_\text{hairdryer}$ is generated and subtracted from the aggregated load curve $L$. 

\paragraph{"Add Lighting".}
As light is not strictly related to activity but rather to  occupancy, this step occurs at the beginning of the "recognize  load profile" process.

\begin{figure}[H]
	\centering
	\includegraphics[scale=\WFscale]{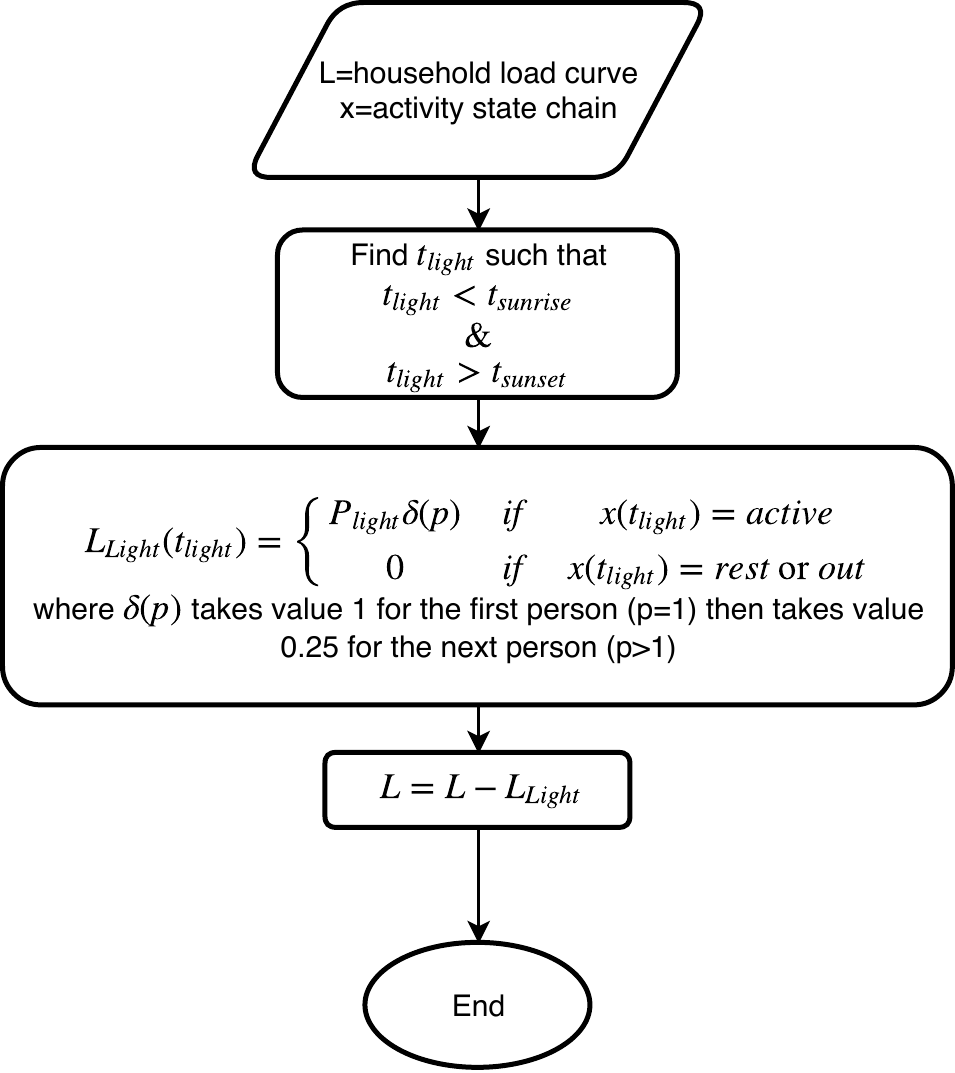}
	\caption{\label{fig:WFaddLight} \bf Add Lighting subprocess.}
\end{figure} 

As the first condition (in figure \ref{fig:WFaddLight}), lighting is assumed to be on only before sunrise and after sunset.
$t_\text{sunrise}$ and $t_\text{sunset}$ are calculated for each day using an approximate equation of time. 
Lighting is assumed to be on only for active people. When people are resting or out of the house, lighting is off. These "no-light"  activities are (with respect to table \ref{tab:ActDev}), \textit{Sleep,Outdoor} and \textit{Work}. In order to avoid overestimating light consumption, a damping variable $\delta(p)$ is introduced. It takes the value of 1 if $p=1$ (first person) then 0.25 for each additional person. 

\begin{figure}[H]
	\centering
	\includegraphics[scale=\WFscale]{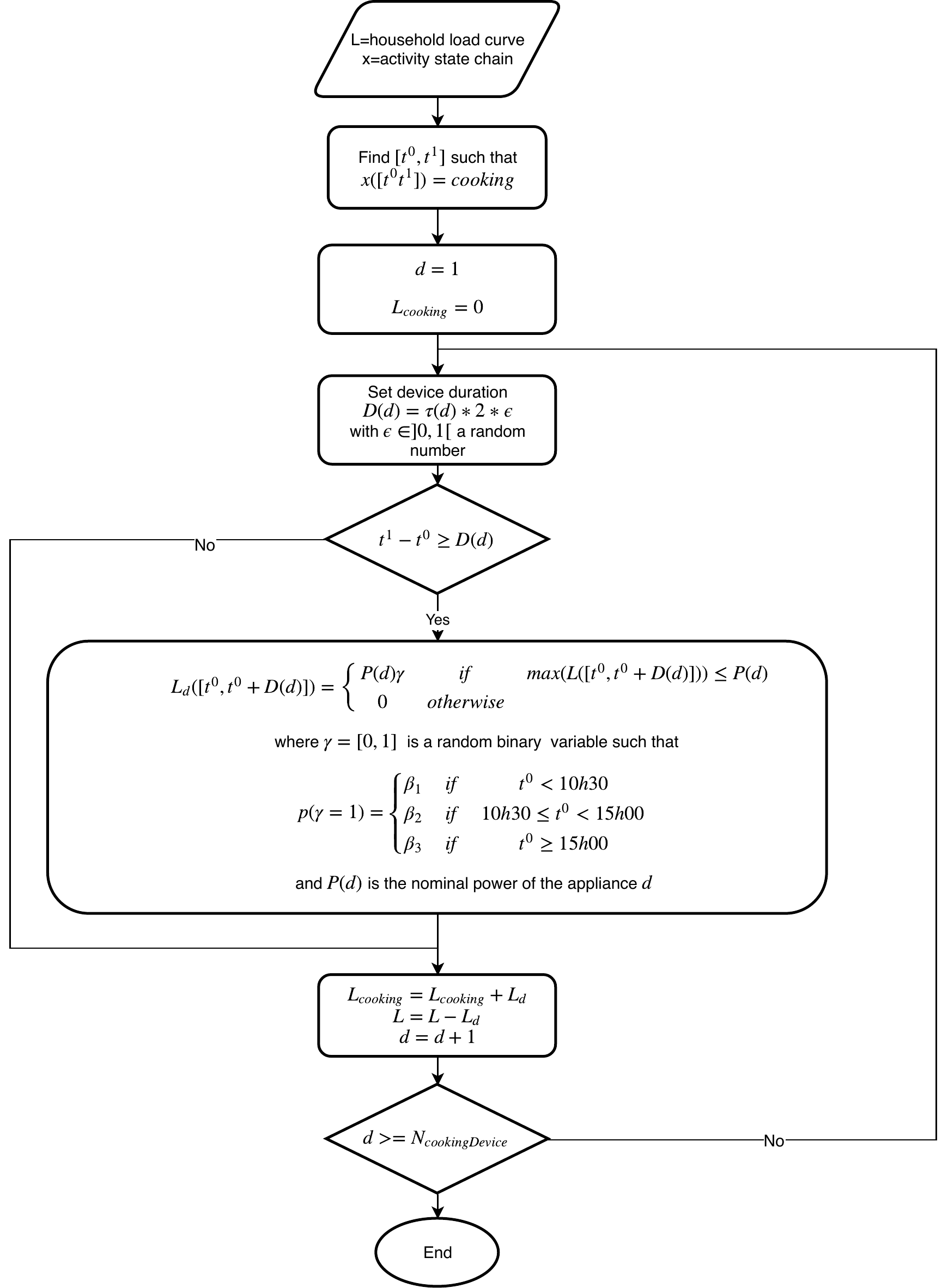}
	\caption{\label{fig:WFcooking} \bf Recognize Cooking subprocess.}
\end{figure}

\paragraph{"Recognize Cooking".}  
As shown in the figure \ref{fig:WFcooking}, the first step is to detect periods corresponding to the possible activities for this category. The cooking-related activities are: \textit{Cook} and \textit{Eat}.  Then, a loop on all possible appliances (see table \ref{tab:ActDev}) is performed. At each iteration, a random appliance usage duration $D$ is selected according to the mean appliance usage duration (reported in table \ref{tab:AppData}). If the length of the activity period is long enough and the power budget is satisfied (condition $max(L([t^0,t^0 +D(d)]) \leq P(d)$, with $P(d)$ the nominal power of the appliance $d$), the power signal for the selected appliance is generated. A Boolean value $\gamma \in [0,1]$ sets the power signal to $0$ with probability $1-\beta_i,i=1..3$ according to three different conditions. These conditions correspond to the actual probability of usage of each particular appliance. In this case it is the probability of usage during breakfast, lunch and dinner time. This is also reported in table \ref{tab:AppData} under footnote \ref{tnote:r1}. 

The final step consists of subtracting the generated appliance power signal $L_d$ from the aggregated load curve $L$.

\paragraph{"Recognize Housekeeping".}
First, the activity period is extracted as previously and a loop on all appliances that might be used by the \textit{Housekeeping} activities is performed. These activities are (according to table \ref{tab:ActDev}) \textit{Clean, Wash dishes} and \textit{Laundry}. The compatibility between the activity duration and the appliance duration is checked before generating the appliance power signal $L_d$. Here again the power budget is checked by comparing the appliance nominal power $P(d)$ with the maximum power of the load $L$. The activation Boolean variable $\gamma$ is used to take into account the probability of usage of each individual appliance. In the general case, probability $\beta_1$ and $\beta_2$ corresponds to the probability of using the appliance if it is the first time the appliance is used or if it was already used during the current day.  In the case of the tumble dryer, it can be used only right after the washing machine with probability $\beta_1$ and only once a day ($\beta_2$=0). The corresponding conditions are given in footnote \ref{tnote:r10} of table \ref{tab:AppData}. 

\begin{figure}[H]
	\centering
	\includegraphics[scale=\WFscale]{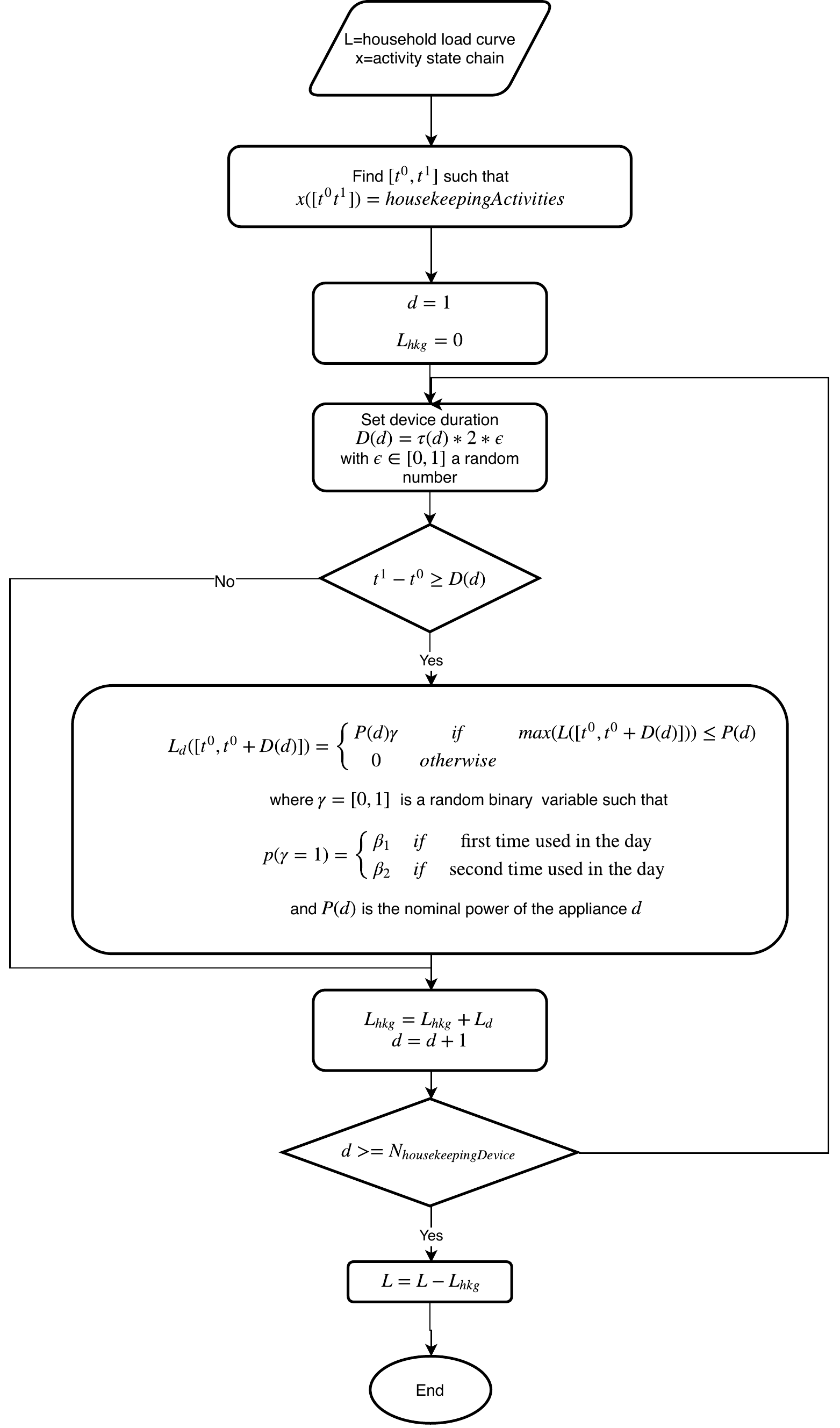}
	\caption{\label{fig:WFhousekeeping} \bf Recognize Housekeeping subprocess.}
\end{figure} 

\begin{figure}[H]
	\centering
	\includegraphics[scale=\WFscale]{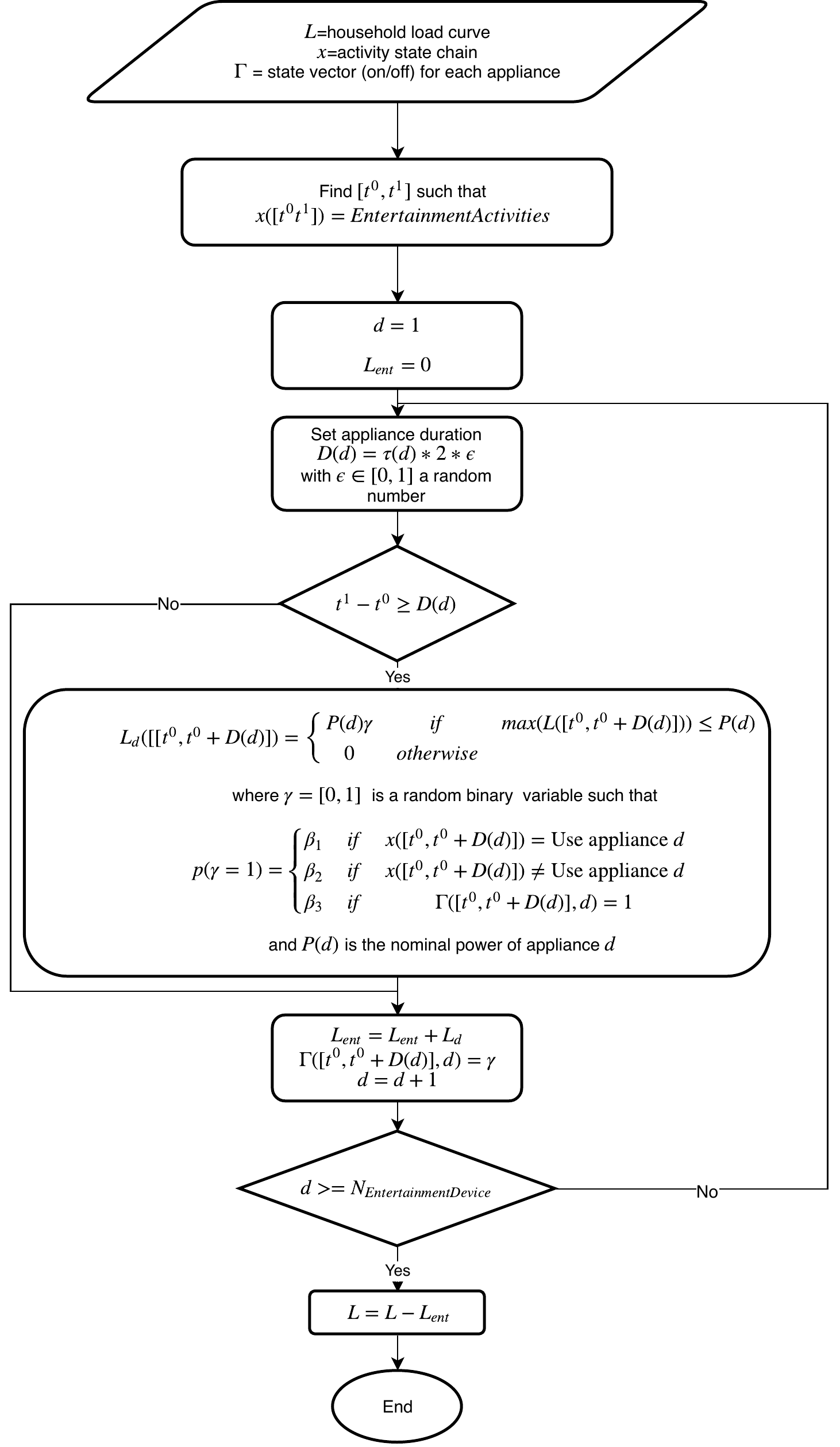}
	\caption{\label{fig:WFentertainment} \bf Add Entertainment  subprocess.}
\end{figure}

\paragraph{"Add Entertainment".}
As entertainment appliances, which are defined in the table \ref{tab:ActDev}, can be used  simultaneously with almost any other activity, some specific adaptation has been made in this subprocess (see figure \ref{fig:WFentertainment}). The probability of usage of a particular appliance is defined according to the condition of whether the activity is specifically to use this appliance. For instance considering TV, if the activity is explicitly \textit{Watch TV}, the usage probability $\beta_1$ is equal to 0.9. This probability is not 1 because this activity could be achieved on other appliances such as a laptop or smartphone. However, if the current activity is \textit{Clean}, the TV can be on in the background with probability $\beta_2=0.1$. Moreover, a state vector $\Gamma$ tracks the on/off state of each appliance. If more than one appliance of the same appliance is in the household inventory and the first appliance is already on ($\Gamma([t^0,t^0+D(d)],d)=1$), the probability of usage of the next appliance is $\beta_3$. Finally, a particular condition is formulated for households that do not own a TV. The activity \textit{Watch TV} can still occur but the TV can be replaced by a PC or laptop with a specific probability of usage (see table \ref{tab:AppData}, footnote \ref{tnote:r5}). At the end of the iteration, the state vector $\Gamma$ is updated with the activation variable $\gamma$. 

\begin{figure}[H]
	\centering
	\includegraphics[scale=\WFscale]{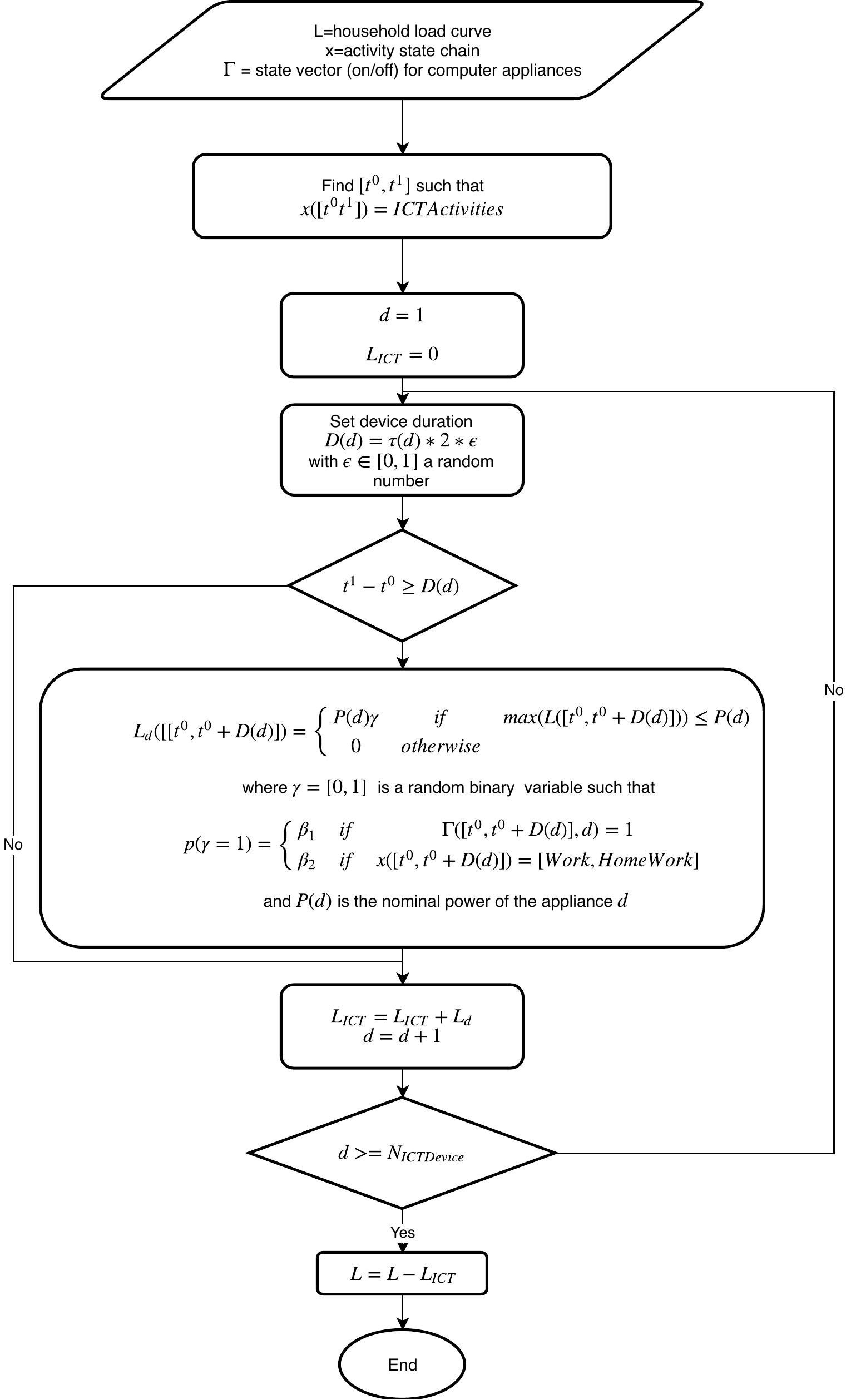}
	\caption{\label{fig:WFICT} \bf Add ICT subprocess.}
\end{figure}

\paragraph{"Add ICT".}
The subprocess to add ICT, as seen in figure \ref{fig:WFICT}, is similar to the previous subprocess with one major exception. The probability of usage depends on either the state $\Gamma$ of the computer appliances (PC and laptop) or  whether the activity is \textit{Work} or \textit{Homework}.  

\section{Household information}
\label{APP:HsldInfo}
The household information is the input of the algorithm. It can be classified into three different categories by order of importance. First is the household composition, namely the total number of people, the number of children below 10 years old, and the number of teenagers. Second is the employment state. Third is the age group of each person, which can actually be deduced from the first categories. Fourth is optional but useful: information about household habits. Any information in this category modifies slightly either the activity probability, or the device usage probability as described in the table note of table \ref{tab:AppData}. Finally, any house information can be used. In this implementation of the methodology, any electrical heating for space or hot water is not considered. Hence it is assumed that no household owns any such device. The household information for each dataset is summarized in table \ref{tab:HsldInfo}. 

In addition, an appliance list has to be provided for each household. The number of appliances owned by each household of the test dataset is summarized in table \ref{tab:HsldApp}. Although this information can be assumed, in the context of this paper, it has been reported as closely as possible to the appliance list provided by the metadata of each dataset. 
\begin{table}[H]
\centering
	\caption{\label{tab:HsldInfo} \bf Household information, assumed or retrieved from dataset metadata.}
\begin{tabular}{lccc}
\toprule
& \textbf{\textsc{ECO}}           & \textbf{\textsc{SMARTENERGY.KOM}} & \textbf{\textsc{UK-DALE}}      \\
\midrule
\textbf{Household composition}              &               &                 &               \\
Total number of people                              & 2             & 1               & 4             \\
number of children ($<$10 y.o)              & 0             & 0               & 0             \\
number of teenagers                                 & 0             & 0               & 2             \\
\midrule
\textbf{Employment state}                            &               &                 &               \\
Person 1                                            & full-time     & full-time       & full-time     \\
Person 2                                            & full-time     &                 & full-time     \\
Person 3                                            &               &                 & student       \\
Person 4                                            &               &                 & student       \\
\midrule
\textbf{Age}                                        &               &                 &               \\
Person 1                                            & senior active & senior active   & senior active \\
Person 2                                            & senior active &                 & senior active \\
Person 3                                            &               &                 & teenager      \\
Person 4                                            &               &                 & teenager      \\
\midrule
\textbf{Houshold habits}                            &               &                 &               \\
Usage of the washing machine per week               & 0             & 1               & 1             \\
Usage of the tumble dryer per week& 0             & 0               & 0             \\
Usage of the dishwasher per week                    & 0             & 0               & 4     \\
Usage of the computer                               & normal        & normal          & occasional    \\
Usage of the TV                            & occasional    & occasional      & occasional    \\
Usage of the stereo                                   & high          & normal          & high          \\
Usage of gaming console                                    & high          & normal          & high          \\
\# lunch at home per week                              & 7             & 7               & 7             \\
\# dinner at home per week                             & 7             & 7               & 7 \\
\midrule
\textbf{House information}                          &               &                 &               \\

Electrical heating                                  & No            & No              & No            \\
\end{tabular}
\end{table}

\begin{table}[H]
\centering
	\caption{\label{tab:HsldApp} \bf Number of appliances owned for each household.}
\begin{tabular}{lccc}
\toprule
& \textbf{\textsc{ECO}}           & \textbf{\textsc{SMARTENERGY.KOM}} & \textbf{\textsc{UK-DALE}}      \\
\midrule
coffee maker                                        & 1             & 1               & 1             \\
microwave                                           & 0             & 0               & 1             \\
kettle                                              & 1             & 1               & 1             \\
oven                                                & 1             & 1               & 1             \\
stove                                               & 1             & 1               & 1             \\
TV                                                  & 1             & 1               & 1             \\
PC                                                  & 1             & 1               & 1             \\
tablet                                              & 0             & 0               & 0             \\
stereo                                              & 1             & 1               & 1             \\
DVD                                                 & 0             & 1               & 1             \\
gaming console                                      & 1             & 1               & 0             \\
TV box                                              & 0             & 0               & 1             \\
laptop                                              & 0             & 0               & 0              \\
fridge (with Freezer)                               & 1             & 0               & 1             \\
freezer                                             & 0             & 1               & 0             \\
fridge(without freezer)                             & 0             & 0               & 0             \\
hairdryer                                           & 0             & 0               & 1             \\
boiler                                              & 0             & 0               & 0             \\
heat-pump                                           & 0             & 0               & 0             \\
washing machine                                     & 0             & 0               & 1             \\
tumble dryer                                        & 0             & 0               & 0             \\
dishwasher                                          & 0             & 0               & 1             \\
vacuum                                              & 0             & 0               & 1             \\
printer                                             & 1             & 1               & 1             \\
lighting                                           & 1             & 1               & 1             \\
modem                                               & 1             & 1               & 1             \\  
\end{tabular}
\end{table}


\clearpage
\bibliographystyle{model1-num-names}
\bibliography{bibliography}

\begin{thebibliography}{56}
\expandafter\ifx\csname natexlab\endcsname\relax\def\natexlab#1{#1}\fi
\providecommand{\bibinfo}[2]{#2}
\ifx\xfnm\relax \def\xfnm[#1]{\unskip,\space#1}\fi
\bibitem[{Strbac(2008)}]{Strbac2008}
\bibinfo{author}{G.~Strbac},
\newblock \bibinfo{title}{{Demand side management: Benefits and challenges}},
\newblock \bibinfo{journal}{Energy Policy} \bibinfo{volume}{36}
  (\bibinfo{year}{2008}) \bibinfo{pages}{4419--4426}.
\bibitem[{Hart(1992)}]{Hart1992}
\bibinfo{author}{G.~W. Hart},
\newblock \bibinfo{title}{{Nonintrusive appliance load monitoring}},
\newblock \bibinfo{journal}{Proceedings of the IEEE} \bibinfo{volume}{80}
  (\bibinfo{year}{1992}) \bibinfo{pages}{1870--1891}.
\bibitem[{Esa et~al.(2016)Esa, Abdullah, and Hassan}]{Esa2016}
\bibinfo{author}{N.~F. Esa}, \bibinfo{author}{M.~P. Abdullah},
  \bibinfo{author}{M.~Y. Hassan},
\newblock \bibinfo{title}{{A review disaggregation method in Non-intrusive
  Appliance Load Monitoring}},
\newblock \bibinfo{journal}{Renewable and Sustainable Energy Reviews}
  \bibinfo{volume}{66} (\bibinfo{year}{2016}) \bibinfo{pages}{163--173}.
\bibitem[{Liang et~al.(2010)Liang, Ng, Kendall, and Cheng}]{Liang2010}
\bibinfo{author}{J.~Liang}, \bibinfo{author}{S.~K.~K. Ng},
  \bibinfo{author}{G.~Kendall}, \bibinfo{author}{J.~W.~M. Cheng},
\newblock \bibinfo{title}{{Load Signature Study—Part I: Basic Concept,
  Structure, and Methodology}},
\newblock \bibinfo{journal}{IEEE Transactions on Power Delivery}
  \bibinfo{volume}{25} (\bibinfo{year}{2010}) \bibinfo{pages}{551--560}.
\bibitem[{Wang et~al.(2018)Wang, Chen, Hong, and Kang}]{Wang2018}
\bibinfo{author}{Y.~Wang}, \bibinfo{author}{Q.~Chen},
  \bibinfo{author}{T.~Hong}, \bibinfo{author}{C.~Kang},
\newblock \bibinfo{title}{{Review of Smart Meter Data Analytics: Applications,
  Methodologies, and Challenges}},
\newblock \bibinfo{journal}{IEEE Transactions on Smart Grid}
  (\bibinfo{year}{2018}) \bibinfo{pages}{1--1}.
\bibitem[{Zoha et~al.(2012)Zoha, Gluhak, Imran, and Rajasegarar}]{Zoha2012}
\bibinfo{author}{A.~Zoha}, \bibinfo{author}{A.~Gluhak},
  \bibinfo{author}{M.~Imran}, \bibinfo{author}{S.~Rajasegarar},
\newblock \bibinfo{title}{{Non-Intrusive Load Monitoring Approaches for
  Disaggregated Energy Sensing: A Survey}},
\newblock \bibinfo{journal}{Sensors} \bibinfo{volume}{12}
  (\bibinfo{year}{2012}) \bibinfo{pages}{16838--16866}.
\bibitem[{Birt et~al.(2012)Birt, Newsham, Beausoleil-Morrison, Armstrong,
  Saldanha, and Rowlands}]{Birt2012}
\bibinfo{author}{B.~J. Birt}, \bibinfo{author}{G.~R. Newsham},
  \bibinfo{author}{I.~Beausoleil-Morrison}, \bibinfo{author}{M.~M. Armstrong},
  \bibinfo{author}{N.~Saldanha}, \bibinfo{author}{I.~H. Rowlands},
\newblock \bibinfo{title}{{Disaggregating categories of electrical energy
  end-use from whole-house hourly data}},
\newblock \bibinfo{journal}{Energy and Buildings} \bibinfo{volume}{50}
  (\bibinfo{year}{2012}) \bibinfo{pages}{93--102}.
\bibitem[{Zhao et~al.(2018)Zhao, Stankovic, and Stankovic}]{1hourStankovic}
\bibinfo{author}{B.~Zhao}, \bibinfo{author}{L.~Stankovic},
  \bibinfo{author}{V.~Stankovic},
\newblock \bibinfo{title}{Electricity usage profile disaggregation of hourly
  smart meter data},
\newblock in: \bibinfo{booktitle}{4th International Workshop on Non-Intrusive
  Load Monitoring}.
\bibitem[{Batra et~al.(2016)Batra, Singh, and Whitehouse}]{Batra2016}
\bibinfo{author}{N.~Batra}, \bibinfo{author}{A.~Singh},
  \bibinfo{author}{K.~Whitehouse},
\newblock \bibinfo{title}{{Gemello}},
\newblock in: \bibinfo{booktitle}{Proceedings of the 22nd ACM SIGKDD
  International Conference on Knowledge Discovery and Data Mining - KDD '16},
  \bibinfo{publisher}{ACM Press}, \bibinfo{address}{New York, New York, USA},
  \bibinfo{year}{2016}, pp. \bibinfo{pages}{431--440}.
\bibitem[{Faustine et~al.(2017)Faustine, Mvungi, Kaijage, and
  Michael}]{Faustine2017}
\bibinfo{author}{A.~Faustine}, \bibinfo{author}{N.~H. Mvungi},
  \bibinfo{author}{S.~Kaijage}, \bibinfo{author}{K.~Michael},
\newblock \bibinfo{title}{{A Survey on Non-Intrusive Load Monitoring Methodies
  and Techniques for Energy Disaggregation Problem}},
\newblock \bibinfo{journal}{arXiv:1703.00785 [cs]}  (\bibinfo{year}{2017}).
\bibitem[{Wang(2003)}]{Wang2003}
\bibinfo{author}{S.-C. Wang},
\newblock \bibinfo{title}{{Artificial Neural Network}},
\newblock in: \bibinfo{booktitle}{Interdisciplinary Computing in Java
  Programming}, \bibinfo{publisher}{Springer US}, \bibinfo{address}{Boston,
  MA}, \bibinfo{year}{2003}, pp. \bibinfo{pages}{81--100}.
\bibitem[{Ruzzelli et~al.(2010)Ruzzelli, Nicolas, Schoofs, and
  O'Hare}]{Ruzzelli2010}
\bibinfo{author}{A.~G. Ruzzelli}, \bibinfo{author}{C.~Nicolas},
  \bibinfo{author}{A.~Schoofs}, \bibinfo{author}{G.~M.~P. O'Hare},
\newblock \bibinfo{title}{{Real-Time Recognition and Profiling of Appliances
  through a Single Electricity Sensor}},
\newblock in: \bibinfo{booktitle}{2010 7th Annual IEEE Communications Society
  Conference on Sensor, Mesh and Ad Hoc Communications and Networks (SECON)},
  \bibinfo{publisher}{IEEE}, \bibinfo{year}{2010}, pp. \bibinfo{pages}{1--9}.
\bibitem[{Kelly and Knottenbelt(2015)}]{kelly_neural_2015}
\bibinfo{author}{J.~Kelly}, \bibinfo{author}{W.~Knottenbelt},
\newblock \bibinfo{title}{Neural {NILM}: {Deep} neural networks applied to
  energy disaggregation},
\newblock in: \bibinfo{booktitle}{Proceedings of the 2nd {ACM} {International}
  {Conference} on {Embedded} {Systems} for {Energy}-{Efficient} {Built}
  {Environments}}, \bibinfo{publisher}{ACM}, \bibinfo{year}{2015}, pp.
  \bibinfo{pages}{55--64}.
\bibitem[{Biansoongnern and Plangklang(2016)}]{Biansoongnern2016}
\bibinfo{author}{S.~Biansoongnern}, \bibinfo{author}{B.~Plangklang},
\newblock \bibinfo{title}{{Nonintrusive load monitoring (NILM) using an
  Artificial Neural Network in embedded system with low sampling rate}},
\newblock in: \bibinfo{booktitle}{2016 13th International Conference on
  Electrical Engineering/Electronics, Computer, Telecommunications and
  Information Technology, ECTI-CON 2016}, pp. \bibinfo{pages}{1--4}.
\bibitem[{Liao et~al.(2014)Liao, Elafoudi, Stankovic, and Stankovic}]{Liao2014}
\bibinfo{author}{J.~Liao}, \bibinfo{author}{G.~Elafoudi},
  \bibinfo{author}{L.~Stankovic}, \bibinfo{author}{V.~Stankovic},
\newblock \bibinfo{title}{{Non-intrusive appliance load monitoring using
  low-resolution smart meter data}},
\newblock in: \bibinfo{booktitle}{2014 IEEE International Conference on Smart
  Grid Communications (SmartGridComm)}, \bibinfo{publisher}{IEEE},
  \bibinfo{year}{2014}, pp. \bibinfo{pages}{535--540}.
\bibitem[{Stankovic et~al.(2016)Stankovic, Stankovic, Liao, and
  Wilson}]{Stankovic2016}
\bibinfo{author}{L.~Stankovic}, \bibinfo{author}{V.~Stankovic},
  \bibinfo{author}{J.~Liao}, \bibinfo{author}{C.~Wilson},
\newblock \bibinfo{title}{{Measuring the energy intensity of domestic
  activities from smart meter data}},
\newblock \bibinfo{journal}{Applied Energy} \bibinfo{volume}{183}
  (\bibinfo{year}{2016}) \bibinfo{pages}{1565--1580}.
\bibitem[{Mairal et~al.(2009)Mairal, Bach, Ponce, and Sapiro}]{Mairal}
\bibinfo{author}{J.~Mairal}, \bibinfo{author}{F.~Bach},
  \bibinfo{author}{J.~Ponce}, \bibinfo{author}{G.~Sapiro},
\newblock \bibinfo{title}{{Online dictionary learning for sparse coding}},
\newblock in: \bibinfo{booktitle}{Proceedings of the 26th Annual International
  Conference on Machine Learning}, \bibinfo{publisher}{ACM Press},
  \bibinfo{address}{New York, New York, USA}, \bibinfo{year}{2009}, pp.
  \bibinfo{pages}{689--696}.
\bibitem[{Kolter et~al.(2010)Kolter, Batra, and Ng}]{Kolter2010}
\bibinfo{author}{J.~Z. Kolter}, \bibinfo{author}{S.~Batra},
  \bibinfo{author}{A.~Y. Ng},
\newblock \bibinfo{title}{{Energy disaggregation via discriminative sparse
  coding}},
\newblock in: \bibinfo{booktitle}{Advances in Neural Information Processing
  Systems}, pp. \bibinfo{pages}{1153--1161}.
\bibitem[{Elhamifar and Sastry(2015)}]{Elhamifar2015}
\bibinfo{author}{E.~Elhamifar}, \bibinfo{author}{S.~Sastry},
\newblock \bibinfo{title}{{Energy Disaggregation via Learning Powerlets and
  Sparse Coding.}},
\newblock in: \bibinfo{booktitle}{Proceedings of the Twenty-Ninth AAAI
  Conference on Artificial Intelligence Pattern}, pp.
  \bibinfo{pages}{629--635}.
\bibitem[{Singh and Majumdar(2017)}]{Singh2017}
\bibinfo{author}{S.~Singh}, \bibinfo{author}{A.~Majumdar},
\newblock \bibinfo{title}{{Deep Sparse Coding for Non-Intrusive Load
  Monitoring}},
\newblock \bibinfo{journal}{IEEE Transactions on Smart Grid}
  (\bibinfo{year}{2017}) \bibinfo{pages}{1--1}.
\bibitem[{Rahimpour et~al.(2017)Rahimpour, Qi, Fugate, and
  Kuruganti}]{Rahimpour2017}
\bibinfo{author}{A.~Rahimpour}, \bibinfo{author}{H.~Qi},
  \bibinfo{author}{D.~Fugate}, \bibinfo{author}{T.~Kuruganti},
\newblock \bibinfo{title}{{Non-Intrusive Energy Disaggregation Using
  Non-negative Matrix Factorization with Sum-to-k Constraint}},
\newblock \bibinfo{journal}{IEEE Transactions on Power Systems}
  \bibinfo{volume}{PP} (\bibinfo{year}{2017}) \bibinfo{pages}{1--1}.
\bibitem[{Bonfigli et~al.(2015)Bonfigli, Squartini, Fagiani, and
  Piazza}]{Bonfigli2015}
\bibinfo{author}{R.~Bonfigli}, \bibinfo{author}{S.~Squartini},
  \bibinfo{author}{M.~Fagiani}, \bibinfo{author}{F.~Piazza},
\newblock \bibinfo{title}{{Unsupervised algorithms for non-intrusive load
  monitoring: An up-to-date overview}},
\newblock in: \bibinfo{booktitle}{2015 IEEE 15th International Conference on
  Environment and Electrical Engineering (EEEIC)}, \bibinfo{publisher}{IEEE},
  \bibinfo{year}{2015}, pp. \bibinfo{pages}{1175--1180}.
\bibitem[{Zoha et~al.(2013)Zoha, Gluhak, Nati, and Imran}]{Zoha2013}
\bibinfo{author}{A.~Zoha}, \bibinfo{author}{A.~Gluhak},
  \bibinfo{author}{M.~Nati}, \bibinfo{author}{M.~A. Imran},
\newblock \bibinfo{title}{{Low-power appliance monitoring using Factorial
  Hidden Markov Models}},
\newblock in: \bibinfo{booktitle}{2013 IEEE Eighth International Conference on
  Intelligent Sensors, Sensor Networks and Information Processing},
  \bibinfo{publisher}{IEEE}, \bibinfo{year}{2013}, pp.
  \bibinfo{pages}{527--532}.
\bibitem[{Kim et~al.(2011)Kim, Marwah, Arlitt, Lyon, and Han}]{Kim}
\bibinfo{author}{H.~Kim}, \bibinfo{author}{M.~Marwah},
  \bibinfo{author}{M.~Arlitt}, \bibinfo{author}{G.~Lyon},
  \bibinfo{author}{J.~Han},
\newblock \bibinfo{title}{{Unsupervised Disaggregation of Low Frequency Power
  Measurements}},
\newblock in: \bibinfo{booktitle}{Proceedings of the 2011 SIAM International
  Conference on Data Mining}, \bibinfo{publisher}{Society for Industrial and
  Applied Mathematics}, \bibinfo{address}{Philadelphia, PA},
  \bibinfo{year}{2011}, pp. \bibinfo{pages}{747--758}.
\bibitem[{Batra et~al.(2014)Batra, Kelly, Parson, Dutta, Knottenbelt, Rogers,
  Singh, and Srivastava}]{Batra2014a}
\bibinfo{author}{N.~Batra}, \bibinfo{author}{J.~Kelly},
  \bibinfo{author}{O.~Parson}, \bibinfo{author}{H.~Dutta},
  \bibinfo{author}{W.~Knottenbelt}, \bibinfo{author}{A.~Rogers},
  \bibinfo{author}{A.~Singh}, \bibinfo{author}{M.~Srivastava},
\newblock \bibinfo{title}{{NILMTK: An Open Source Toolkit for Non-intrusive
  Load Monitoring}},
\newblock \bibinfo{journal}{Sensors (Switzerland)} \bibinfo{volume}{12}
  (\bibinfo{year}{2014}) \bibinfo{pages}{16838--16866}.
\bibitem[{Bonfigli et~al.(2017)Bonfigli, Principi, Fagiani, Severini,
  Squartini, and Piazza}]{Bonfigli2017}
\bibinfo{author}{R.~Bonfigli}, \bibinfo{author}{E.~Principi},
  \bibinfo{author}{M.~Fagiani}, \bibinfo{author}{M.~Severini},
  \bibinfo{author}{S.~Squartini}, \bibinfo{author}{F.~Piazza},
\newblock \bibinfo{title}{{Non-intrusive load monitoring by using active and
  reactive power in additive Factorial Hidden Markov Models}},
\newblock \bibinfo{journal}{Applied Energy} \bibinfo{volume}{208}
  (\bibinfo{year}{2017}) \bibinfo{pages}{1590--1607}.
\bibitem[{Kumar and Chandra(2017)}]{Kumar2017a}
\bibinfo{author}{K.~Kumar}, \bibinfo{author}{M.~G. Chandra},
\newblock \bibinfo{title}{{An intuitive explanation of graph signal
  processing-based electrical load disaggregation}},
\newblock in: \bibinfo{booktitle}{2017 IEEE 13th International Colloquium on
  Signal Processing {\&} its Applications (CSPA)}, \bibinfo{publisher}{IEEE},
  \bibinfo{year}{2017}, pp. \bibinfo{pages}{100--105}.
\bibitem[{He et~al.(2016)He, Stankovic, Liao, and Stankovic}]{He2016}
\bibinfo{author}{K.~He}, \bibinfo{author}{L.~Stankovic},
  \bibinfo{author}{J.~Liao}, \bibinfo{author}{V.~Stankovic},
\newblock \bibinfo{title}{{Non-Intrusive Load Disaggregation using Graph Signal
  Processing}},
\newblock \bibinfo{journal}{IEEE Transactions on Smart Grid}
  \bibinfo{volume}{9} (\bibinfo{year}{2016}) \bibinfo{pages}{1739--1747}.
\bibitem[{Kumar et~al.(2016)Kumar, Sinha, Chandra, and Thokala}]{Kumar2016}
\bibinfo{author}{K.~Kumar}, \bibinfo{author}{R.~Sinha}, \bibinfo{author}{M.~G.
  Chandra}, \bibinfo{author}{N.~K. Thokala},
\newblock \bibinfo{title}{{Data-driven electrical load disaggregation using
  graph signal processing}},
\newblock in: \bibinfo{booktitle}{2016 IEEE Annual India Conference (INDICON)},
  \bibinfo{publisher}{IEEE}, \bibinfo{year}{2016}, pp. \bibinfo{pages}{1--6}.
\bibitem[{Kumar and Chandra(2017)}]{Kumar2017}
\bibinfo{author}{K.~Kumar}, \bibinfo{author}{M.~G. Chandra},
\newblock \bibinfo{title}{{Event and feature based electrical load
  disaggregation using graph signal processing}},
\newblock in: \bibinfo{booktitle}{2017 IEEE 13th International Colloquium on
  Signal Processing {\&} its Applications (CSPA)}, \bibinfo{publisher}{IEEE},
  \bibinfo{year}{2017}, pp. \bibinfo{pages}{168--172}.
\bibitem[{Zhao et~al.(2016)Zhao, Stankovic, and Stankovic}]{Zhao2016}
\bibinfo{author}{B.~Zhao}, \bibinfo{author}{L.~Stankovic},
  \bibinfo{author}{V.~Stankovic},
\newblock \bibinfo{title}{{On a Training-Less Solution for Non-Intrusive
  Appliance Load Monitoring Using Graph Signal Processing}},
\newblock \bibinfo{journal}{IEEE Access} \bibinfo{volume}{4}
  (\bibinfo{year}{2016}) \bibinfo{pages}{1784--1799}.
\bibitem[{{Sociaal en Cultureel
  Planbureau}(2005)}]{SociaalenCultureelPlanbureau2005}
\bibinfo{author}{{Sociaal en Cultureel Planbureau}},
\newblock \bibinfo{title}{{Tijdsbestedingsonderzoek 2005 - TBO 2005}},
\newblock \bibinfo{journal}{DANS}  (\bibinfo{year}{2005}).
\bibitem[{Rabiner(1989)}]{Rabiner1989a}
\bibinfo{author}{L.~Rabiner},
\newblock \bibinfo{title}{{A tutorial on hidden Markov models and selected
  applications in speech recognition}},
\newblock \bibinfo{journal}{Proceedings of the IEEE} \bibinfo{volume}{77}
  (\bibinfo{year}{1989}) \bibinfo{pages}{257--286}.
\bibitem[{Gershuny and Sullivan(2017)}]{Gershuny2017}
\bibinfo{author}{J.~Gershuny}, \bibinfo{author}{O.~Sullivan},
\newblock \bibinfo{title}{{United Kingdom Time Use Survey, 2014-2015}},
\newblock \bibinfo{journal}{Centre for Time Use Research, University of Oxford,
  UK Data Service}  (\bibinfo{year}{2017}).
\bibitem[{Billauer(2012)}]{Billauer2012}
\bibinfo{author}{E.~Billauer}, \bibinfo{title}{{peakdet: Peak detection using
  MATLAB (non-derivative local extremum, maximum, minimum)}},
  \bibinfo{year}{2012}.
\bibitem[{Kolter and Johnson(2011)}]{Kolter}
\bibinfo{author}{J.~Z. Kolter}, \bibinfo{author}{M.~J. Johnson},
\newblock \bibinfo{title}{{REDD: A Public Data Set for Energy Disaggregation
  Research}},
\newblock in: \bibinfo{booktitle}{Workshop on Data Mining Applications in
  Sustainability (SIGKDD), San Diego, CA}, volume~\bibinfo{volume}{25}, pp.
  \bibinfo{pages}{59--62}.
\bibitem[{Stankovic et~al.(2014)Stankovic, Liao, and Stankovic}]{Stankovic2014}
\bibinfo{author}{V.~Stankovic}, \bibinfo{author}{J.~Liao},
  \bibinfo{author}{L.~Stankovic},
\newblock \bibinfo{title}{{A graph-based signal processing approach for
  low-rate energy disaggregation}},
\newblock in: \bibinfo{booktitle}{2014 IEEE Symposium on Computational
  Intelligence for Engineering Solutions (CIES)}, \bibinfo{publisher}{IEEE},
  \bibinfo{year}{2014}, pp. \bibinfo{pages}{81--87}.
\bibitem[{Perraudin et~al.(2014)Perraudin, Paratte, Shuman, Martin, Kalofolias,
  Vandergheynst, and Hammond}]{perraudin2014gspbox}
\bibinfo{author}{N.~Perraudin}, \bibinfo{author}{J.~Paratte},
  \bibinfo{author}{D.~Shuman}, \bibinfo{author}{L.~Martin},
  \bibinfo{author}{V.~Kalofolias}, \bibinfo{author}{P.~Vandergheynst},
  \bibinfo{author}{D.~K. Hammond},
\newblock \bibinfo{title}{{GSPBOX: A toolbox for signal processing on graphs}},
\newblock \bibinfo{journal}{ArXiv e-prints}  (\bibinfo{year}{2014}).
\bibitem[{Leijonmarck(2015)}]{Leijonmarck2015}
\bibinfo{author}{E.~Leijonmarck}, \bibinfo{title}{Exploiting Temporal
  Difference for Energy Disaggregation via Discriminative Sparse Coding},
  Master's thesis, KTH, Mathematical Statistics, \bibinfo{year}{2015}.
\bibitem[{Yu et~al.(2016)Yu, Mirowski, and Ho}]{Yu2016}
\bibinfo{author}{C.-N. Yu}, \bibinfo{author}{P.~Mirowski},
  \bibinfo{author}{T.~K. Ho},
\newblock \bibinfo{title}{{A Sparse Coding Approach to Household Electricity
  Demand Forecasting in Smart Grids}},
\newblock \bibinfo{journal}{IEEE Transactions on Smart Grid}
  (\bibinfo{year}{2016}) \bibinfo{pages}{1--11}.
\bibitem[{Hoyer(2012)}]{Hoyer}
\bibinfo{author}{P.~Hoyer},
\newblock \bibinfo{title}{{Non-negative sparse coding}},
\newblock in: \bibinfo{booktitle}{Proceedings of the 12th IEEE Workshop on
  Neural Networks for Signal Processing}, \bibinfo{publisher}{IEEE},
  \bibinfo{year}{2012}, pp. \bibinfo{pages}{557--565}.
\bibitem[{Pyt(2018)}]{Python}
\bibinfo{title}{{Python.org}}, \bibinfo{year}{2018}.
\bibitem[{Beckel et~al.(2014)Beckel, Kleiminger, Cicchetti, Staake, and
  Santini}]{Beckel2014}
\bibinfo{author}{C.~Beckel}, \bibinfo{author}{W.~Kleiminger},
  \bibinfo{author}{R.~Cicchetti}, \bibinfo{author}{T.~Staake},
  \bibinfo{author}{S.~Santini},
\newblock \bibinfo{title}{{The ECO data set and the performance of
  non-intrusive load monitoring algorithms}},
\newblock in: \bibinfo{booktitle}{Proceedings of the 1st ACM Conference on
  Embedded Systems for Energy-Efficient Buildings - BuildSys '14}, pp.
  \bibinfo{pages}{80--89}.
\bibitem[{Alhamoud et~al.(2014)Alhamoud, Ruettiger, Reinhardt, Englert,
  Burgstahler, Bohnstedt, Gottron, and Steinmetz}]{Alhamoud2014}
\bibinfo{author}{A.~Alhamoud}, \bibinfo{author}{F.~Ruettiger},
  \bibinfo{author}{A.~Reinhardt}, \bibinfo{author}{F.~Englert},
  \bibinfo{author}{D.~Burgstahler}, \bibinfo{author}{D.~Bohnstedt},
  \bibinfo{author}{C.~Gottron}, \bibinfo{author}{R.~Steinmetz},
\newblock \bibinfo{title}{{SMARTENERGY.KOM: An intelligent system for energy
  saving in smart home}},
\newblock in: \bibinfo{booktitle}{39th Annual IEEE Conference on Local Computer
  Networks Workshops}, \bibinfo{publisher}{IEEE}, \bibinfo{year}{2014}, pp.
  \bibinfo{pages}{685--692}.
\bibitem[{Kelly and Knottenbelt(2015)}]{Kelly}
\bibinfo{author}{J.~Kelly}, \bibinfo{author}{W.~Knottenbelt},
\newblock \bibinfo{title}{{The UK-DALE dataset, domestic appliance-level
  electricity demand and whole-house demand from five UK homes}},
\newblock \bibinfo{journal}{Scientific Data} \bibinfo{volume}{2}
  (\bibinfo{year}{2015}) \bibinfo{pages}{150007}.
\bibitem[{Beckel et~al.(2014)Beckel, Sadamori, Staake, and
  Santini}]{Beckel2014a}
\bibinfo{author}{C.~Beckel}, \bibinfo{author}{L.~Sadamori},
  \bibinfo{author}{T.~Staake}, \bibinfo{author}{S.~Santini},
\newblock \bibinfo{title}{{Revealing household characteristics from smart meter
  data}},
\newblock \bibinfo{journal}{Energy} \bibinfo{volume}{78} (\bibinfo{year}{2014})
  \bibinfo{pages}{397--410}.
\bibitem[{Alhamoud et~al.(2015)Alhamoud, Xu, Englert, Reinhardt, Scholl,
  Boehnstedt, and Steinmetz}]{Alhamoud2015}
\bibinfo{author}{A.~Alhamoud}, \bibinfo{author}{P.~Xu},
  \bibinfo{author}{F.~Englert}, \bibinfo{author}{A.~Reinhardt},
  \bibinfo{author}{P.~Scholl}, \bibinfo{author}{D.~Boehnstedt},
  \bibinfo{author}{R.~Steinmetz},
\newblock \bibinfo{title}{{Extracting human behavior patterns from
  appliance-level power consumption data}},
\newblock in: \bibinfo{booktitle}{European Conference on Wireless Sensor
  Networks}, \bibinfo{publisher}{Springer}, \bibinfo{year}{2015}, pp.
  \bibinfo{pages}{52--67}.
\bibitem[{Kelly(2017)}]{Kelly2017}
\bibinfo{author}{J.~Kelly}, \bibinfo{title}{{UK Domestic Appliance Level
  Electricity(UK-DALE-2017)-Disaggregated appliance/whole house power}},
  \bibinfo{year}{2017}.
\bibitem[{Liu et~al.(2018)Liu, Akintayo, Jiang, Henze, and Sarkar}]{Liu2018}
\bibinfo{author}{C.~Liu}, \bibinfo{author}{A.~Akintayo},
  \bibinfo{author}{Z.~Jiang}, \bibinfo{author}{G.~P. Henze},
  \bibinfo{author}{S.~Sarkar},
\newblock \bibinfo{title}{{Multivariate exploration of non-intrusive load
  monitoring via spatiotemporal pattern network}},
\newblock \bibinfo{journal}{Applied Energy} \bibinfo{volume}{211}
  (\bibinfo{year}{2018}) \bibinfo{pages}{1106--1122}.
\bibitem[{Kolter et~al.(2012)Kolter, Jaakkola, and Kolter}]{Kolter2012}
\bibinfo{author}{Z.~Kolter}, \bibinfo{author}{T.~Jaakkola},
  \bibinfo{author}{J.~Z. Kolter},
\newblock \bibinfo{title}{{Approximate Inference in Additive Factorial HMMs
  with Application to Energy Disaggregation}},
\newblock \bibinfo{journal}{Proc. Int. Conf. Arti. Intell. Statist.}
  \bibinfo{volume}{XX} (\bibinfo{year}{2012}) \bibinfo{pages}{1472--1482}.
\bibitem[{Dong et~al.(2013)Dong, Wang, and Lu}]{Dong2013}
\bibinfo{author}{H.~Dong}, \bibinfo{author}{B.~Wang}, \bibinfo{author}{C.~T.
  Lu},
\newblock \bibinfo{title}{{Deep sparse coding based recursive disaggregation
  model for water conservation}},
\newblock \bibinfo{journal}{IJCAI International Joint Conference on Artificial
  Intelligence}  (\bibinfo{year}{2013}) \bibinfo{pages}{2804--2810}.
\bibitem[{Parson et~al.(2012)Parson, Ghosh, Weal, and Rogers}]{Parson2012}
\bibinfo{author}{O.~Parson}, \bibinfo{author}{S.~Ghosh},
  \bibinfo{author}{M.~Weal}, \bibinfo{author}{A.~Rogers},
\newblock \bibinfo{title}{{Non-Intrusive Load Monitoring Using Prior Models of
  General Appliance Types}},
\newblock in: \bibinfo{booktitle}{AAAi}.
\bibitem[{Makonin and Popowich(2015)}]{Makonin2015}
\bibinfo{author}{S.~Makonin}, \bibinfo{author}{F.~Popowich},
\newblock \bibinfo{title}{{Nonintrusive load monitoring (NILM) performance
  evaluation}},
\newblock \bibinfo{journal}{Energy Efficiency} \bibinfo{volume}{8}
  (\bibinfo{year}{2015}) \bibinfo{pages}{809--814}.
\bibitem[{Johnson and Willsky(2013)}]{Johnson2013}
\bibinfo{author}{M.~J. Johnson}, \bibinfo{author}{A.~S. Willsky},
\newblock \bibinfo{title}{{Bayesian Nonparametric Hidden Semi-Markov Models}},
\newblock \bibinfo{journal}{Journal of Machine Learning Research}
  \bibinfo{volume}{14} (\bibinfo{year}{2013}) \bibinfo{pages}{673--701}.
\bibitem[{Manivannan et~al.(2017)Manivannan, Najafi, and Rinaldi}]{reviewer2}
\bibinfo{author}{M.~Manivannan}, \bibinfo{author}{B.~Najafi},
  \bibinfo{author}{F.~Rinaldi},
\newblock \bibinfo{title}{{Machine Learning-Based Short-Term Prediction of
  Air-Conditioning Load through Smart Meter Analyticsan intu}},
\newblock \bibinfo{journal}{MDPI Energies} \bibinfo{volume}{10}
  (\bibinfo{year}{2017}).
\bibitem[{Hakell et~al.(2015)Hakell, Fisher, and Hersey}]{Hakell2015}
\bibinfo{author}{B.~Hakell}, \bibinfo{author}{G.~Fisher},
  \bibinfo{author}{A.~Hersey}, \bibinfo{title}{{Setting the Benchmark for Non
  Intrusive Load Monitoring : A Comprehensive Assessment of AMI-based Load
  Disaggregation}}, \bibinfo{type}{Technical Report}, Pecan Street,
  \bibinfo{year}{2015}.

\end{thebibliography}

\end{document}